\documentclass{ar-1col}

\usepackage[usenames,dvipsnames,table]{xcolor}
\usepackage[numbers]{natbib}
\usepackage{url}
\usepackage{amssymb}
\usepackage{amsmath}
\usepackage{float}
\usepackage[inline]{enumitem}
\usepackage{multirow}
\usepackage{array}
\usepackage{caption}
\usepackage{subcaption}
\usepackage{hyperref}
\usepackage{booktabs}
\usepackage{pdflscape}
\usepackage{rotating}
\usepackage{graphics}

\usepackage{tikz}
\usepackage{pgf}
\usepackage{pgfplots}
\usepackage{tikz-3dplot}
\usetikzlibrary{arrows}
\usetikzlibrary{shadows}
\usetikzlibrary{petri}
\usetikzlibrary{automata}
\usetikzlibrary{shapes.geometric}
\usetikzlibrary{shapes.misc}
\usetikzlibrary{positioning}
\usetikzlibrary{fit}
\usetikzlibrary{calc}
\usetikzlibrary{angles}
\usetikzlibrary{quotes}
\usetikzlibrary{fadings}
\usetikzlibrary{decorations.pathreplacing}
\usetikzlibrary{snakes}
\usetikzlibrary{arrows.meta}
\usetikzlibrary{shapes.arrows}
\usetikzlibrary{spy}
\usetikzlibrary{patterns}
\usetikzlibrary{patterns.meta}
\usepgfplotslibrary{groupplots}
\usepgfplotslibrary{fillbetween}
\usepgfplotslibrary{statistics}
\usetikzlibrary{external}
\hypersetup{
    colorlinks=true,
    allcolors=red
}

\newcommand{\comment}[1]{}

\newcommand{\PreserveBackslash}[1]{\let\temp=\\#1\let\\=\temp}
\newcolumntype{C}[1]{>{\PreserveBackslash\centering}p{#1}}
\newcolumntype{R}[1]{>{\PreserveBackslash\raggedleft}p{#1}}
\newcolumntype{L}[1]{>{\PreserveBackslash\raggedright}p{#1}}

\renewcommand\vec{\mathbf} \newcommand{\set}[1]{\mathbb{#1}} \newcommand{\R}{\set{R}}  \newcommand{\dist}[1]{\mathcal{#1}} \usepackage{amsmath}

\DeclareMathOperator*{\argmin}{arg\,min}

\newcommand{\sysstate}{\vec{x}}

\newcommand{\sysinput}{\vec{u}}
\newcommand{\inputdelta}{\psi} \newcommand{\inputdeltalearned}{\hat{\psi}}  \newcommand{\wnoise}{{\vec{w}}}    

\newcommand{\stateconstraints}{\set{X}_{\mathrm{c}} }
\newcommand{\inputconstraints}{\set{U}_{\mathrm{c}}}

\newcommand{\rmpcsysstate}{\vec{z}}

\newcommand{\cost}{J}
\newcommand{\coststep}{l} \newcommand{\costprior}{\bar{\cost}} \newcommand{\costlearned}{\hat{\cost}} \newcommand{\costsoft}{\coststep_\epsilon} 

\newcommand{\dyn}{\vec{f}}
\newcommand{\dynprior}{\bar{\dyn}} \newcommand{\dynlearned}{\hat{\dyn}}

\newcommand{\constraintsingle}{c}
\newcommand{\cstr}{\vec{\constraintsingle}} \newcommand{\cstrprior}{\bar{\cstr}} 

\newcommand{\data}{\mathcal{D}}
\newcommand{\prior}{\mathcal{P}}

\newcommand{\policy}{\boldsymbol{\pi}}

\newcommand{\params}{\boldsymbol{\theta}}

\newcommand{\linearpriorA}{\bar{\vec{A}}}

\newcommand{\linearpriorB}{\bar{\vec{B}}}

\newcommand{\linearunknownA}{\hat{\vec{A}}}

\newcommand{\linearunknownB}{\hat{\vec{B}}}

\newcommand{\lyapunov}{L}
\newcommand{\Hfunc}{H}

\newcommand{\subheading}[1]{\vspace{0.5em}\noindent\textit{\textbf{#1}}}

\usepackage{amssymb}\usepackage{pifont}

\newcounter{margin}
\newcommand*{\marginref}[1]{\hyperref[#1]{M\ref{#1}}}

\newcommand{\aref}[1]{\hyperref[#1]{Appendix~\ref*{#1}}}

\begin{document}

\markboth{Brunke et al.}{Safe Learning in Robotics}

\title{Safe Learning in Robotics:
From Learning-Based Control to Safe Reinforcement Learning}

\author{Lukas Brunke$^{*}$,
Melissa Greeff$^{*}$,
Adam W. Hall$^{*}$,
Zhaocong Yuan$^{*}$,
Siqi Zhou$^{*}$,
Jacopo Panerati,
and Angela P. Schoellig
\affil{$^*$Equal contribution}
\affil{Institute for Aerospace Studies, University of Toronto, Toronto, Ontario, Canada, M3H~5T6; University of Toronto Robotics Institute, Toronto, Ontario, Canada, M5S~1A4; Vector Institute for Artificial Intelligence, Toronto, Ontario, Canada, M5G~1M1;  emails: \texttt{\{firstname.lastname\}@robotics.utias.utoronto.ca} }
}

\begin{abstract}
The last half-decade has seen a steep rise in the number of contributions on safe learning methods for real-world robotic deployments from both the control and reinforcement learning communities. This article provides a concise but holistic review of the recent advances made in using machine learning to achieve safe decision making under uncertainties, with a focus on unifying the language and frameworks used in control theory and reinforcement learning research. Our review includes: learning-based control approaches that safely improve performance by learning the uncertain dynamics, reinforcement learning approaches that encourage safety or robustness, and methods that can formally certify the safety of a learned control policy. As data- and learning-based robot control methods continue to gain traction, researchers must understand when and how to best leverage them in real-world scenarios where safety is imperative, such as when operating in close proximity to humans. We highlight some of the open challenges that will drive the field of robot learning in the coming years, and emphasize the need for realistic physics-based benchmarks to facilitate fair comparisons between control and reinforcement learning approaches.
\end{abstract}

\maketitle

\tableofcontents

\section{INTRODUCTION} \label{sec:introduction}

Robotics researchers strive to design systems that can operate autonomously in increasingly complex scenarios, often in close proximity to humans. Examples include self-driving vehicles~\cite{burnett2021a}, aerial delivery~\cite{boutilier2017}, and the use of mobile manipulators for service tasks~\cite{dong2020a}.
However, the dynamics of these complex applications are often uncertain
or only partially known---for example,  the mass distribution of a carried payload might not be given \emph{a priori}. Uncertainties arise from various sources. For example, the robot dynamics may not be perfectly modeled, sensor measurements may be noisy, and/or the operating environment may not be well-characterized or may include other agents whose dynamics and plans are not known.  

In these real-world applications, robots must \textit{make decisions despite having only partial knowledge of the world}.
In recent years, the research community has multiplied its efforts
to leverage data-based approaches to address this problem. This was motivated in part by the success of machine learning
in other areas such as computer vision and natural language processing. 

A crucial, domain-specific challenge of \textit{learning for robot control} is the need to implement and formally guarantee
safety of the robot's behavior, not only for the optimized policy (or controller, which is essential for the certification of systems
that interact with humans) but also during learning, to avoid costly hardware failures and improve convergence.
Ultimately, these safety guarantees can only be derived from the assumptions and structure captured by the problem formalization.

\begin{figure}
    \centering
    \includegraphics[width=\textwidth]{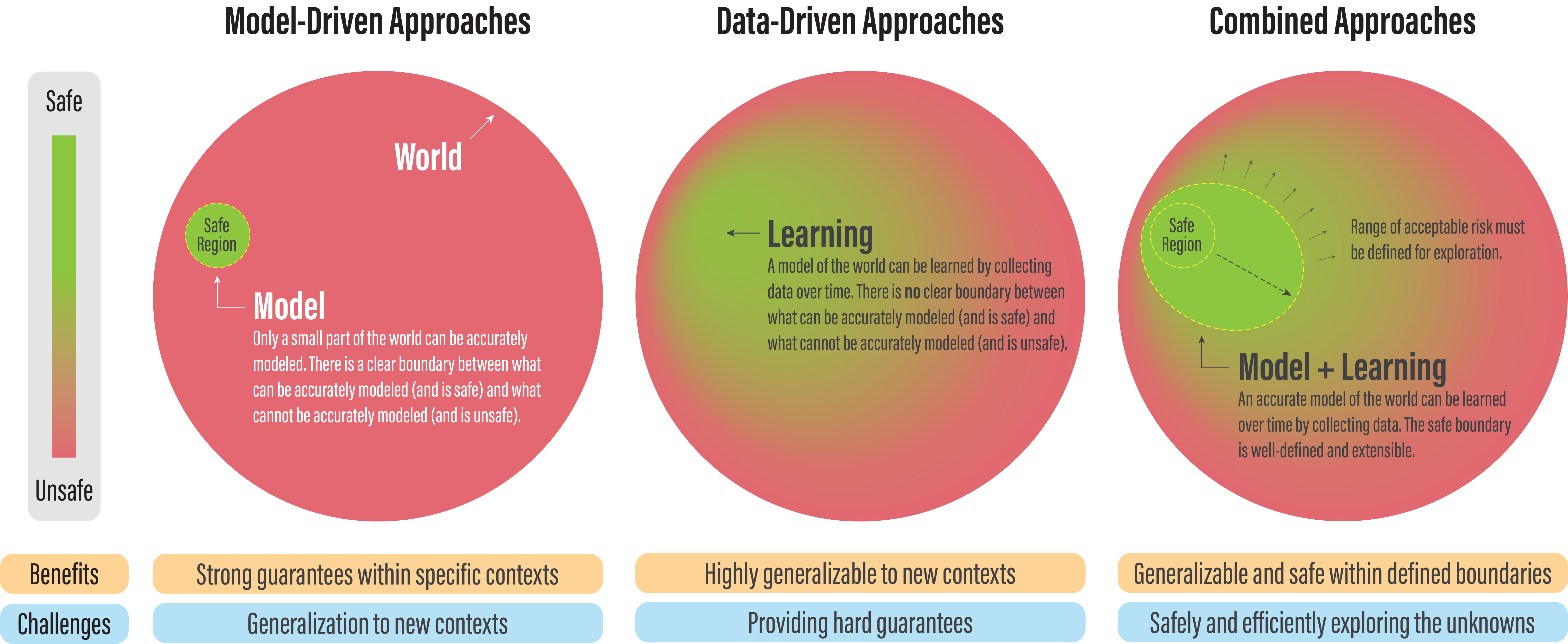}
    \caption{A comparison of model-driven, data-driven, and combined approaches.}
    \label{fig:problem_illustration}
\end{figure}

Both control theory and machine learning---reinforcement learning (RL) in particular---have proposed approaches to tackle this problem. Control theory has traditionally taken a model-driven approach (see \textbf{\autoref{fig:problem_illustration}}): it leverages a given dynamics model and provides guarantees with respect to known operating conditions. RL has traditionally taken a data-driven approach, which makes it highly adaptable to new contexts at the expense of providing formal guarantees. Combining model-driven and data-driven approaches, and  leveraging the advantages of each, is a promising direction for safe learning in robotics. 
The methods we review encourage \textbf{robustness} (by accounting for the worst-case scenarios and taking conservative actions), enable \textbf{adaptation} (by learning from online observations and adapting to unknown situations), and build and leverage \textbf{prediction models} (based on a combination of domain knowledge, real-world data, and high-fidelity simulators).

While control is still the bedrock of all current robot applications, the body of safe RL literature has ballooned from tens
to over a thousand publications in just a few years since its most recent review~\cite{Garcia2015}. Physics-based simulation~\cite{dulacarnold2020a}, which we leverage in our open-source benchmark implementation\footnote{Safe control benchmark suite on GitHub: \url{https://github.com/utiasDSL/safe-control-gym}}, has played an important role in the recent
progress of RL, however, the transfer to real systems remains a research area in itself~\cite{dulac2019a}.

Previous review works focused on specific techniques---for example,
learning-based model predictive control (MPC)~\cite{Hewing2020},
iterative learning control (ILC)~\citep{bristow2006survey,ahn2007iterative},
model-based RL~\cite{Polydoros2017},
data-efficient policy search~\cite{Chatzilygeroudis2020},
imitation learning~\cite{Ravichandar2020},
or the use of RL in robotics~\cite{Kober2013,Recht2019} and in optimal control~\cite{Kiumarsi2018}---without emphasizing the safety aspect. 
Recent surveys on safe learning control focus on either control-theoretic~\cite{Osborne2021} or RL approaches~\cite{Tambon2021}, and do not provide a unifying perspective. 

In this article, we provide a bird's eye view of the most recent work in learning-based control and reinforcement learning that implement safety and provide safety guarantees for robot control. We focus on safe learning control approaches where the data generated by the robot system is used to learn or modify the feedback controller (or policy).
We hope to help shrink the gap between the control and RL communities by creating a common vocabulary and introducing benchmarks for algorithm evaluation that can be leveraged by both~\cite{ray2019a,leike2017a}.
Our target audience are researchers, with either a control or RL background, who are interested in a concise but holistic perspective on the problem of safe learning control.
While we do not cover perception, estimation, planning, or multi-agent systems, we do connect our discussion to these additional challenges and opportunities. 

\section{PRELIMINARIES AND BACKGROUND OF SAFE LEARNING CONTROL} \label{sec:background}

In this review, we are interested in the problem of \textit{safe decision making under uncertainties using machine learning} (i.e., \textit{safe learning control}). Intuitively, in safe learning control, our goal is to allow a robot to fulfil a task while respecting a set of safety constraints 
despite the uncertainties present in the problem. In this section, we define the safe learning control problem (\autoref{subsec:problem_statement}) and provide an overview of how the problem of \textit{decision making under uncertainties} has traditionally been tackled  by the control (\autoref{subsec:control_background}) and RL communities (\autoref{subsec:reinforcement_learning_background}). We highlight the main limitations of these approaches and articulate how novel data-based, safety-focused methods can address these gaps (\autoref{subsec:data_in_safe_learning}).

\subsection{Problem Statement}
\label{subsec:problem_statement}
\begin{figure}\centering
    \includegraphics[draft=false, width=\textwidth]{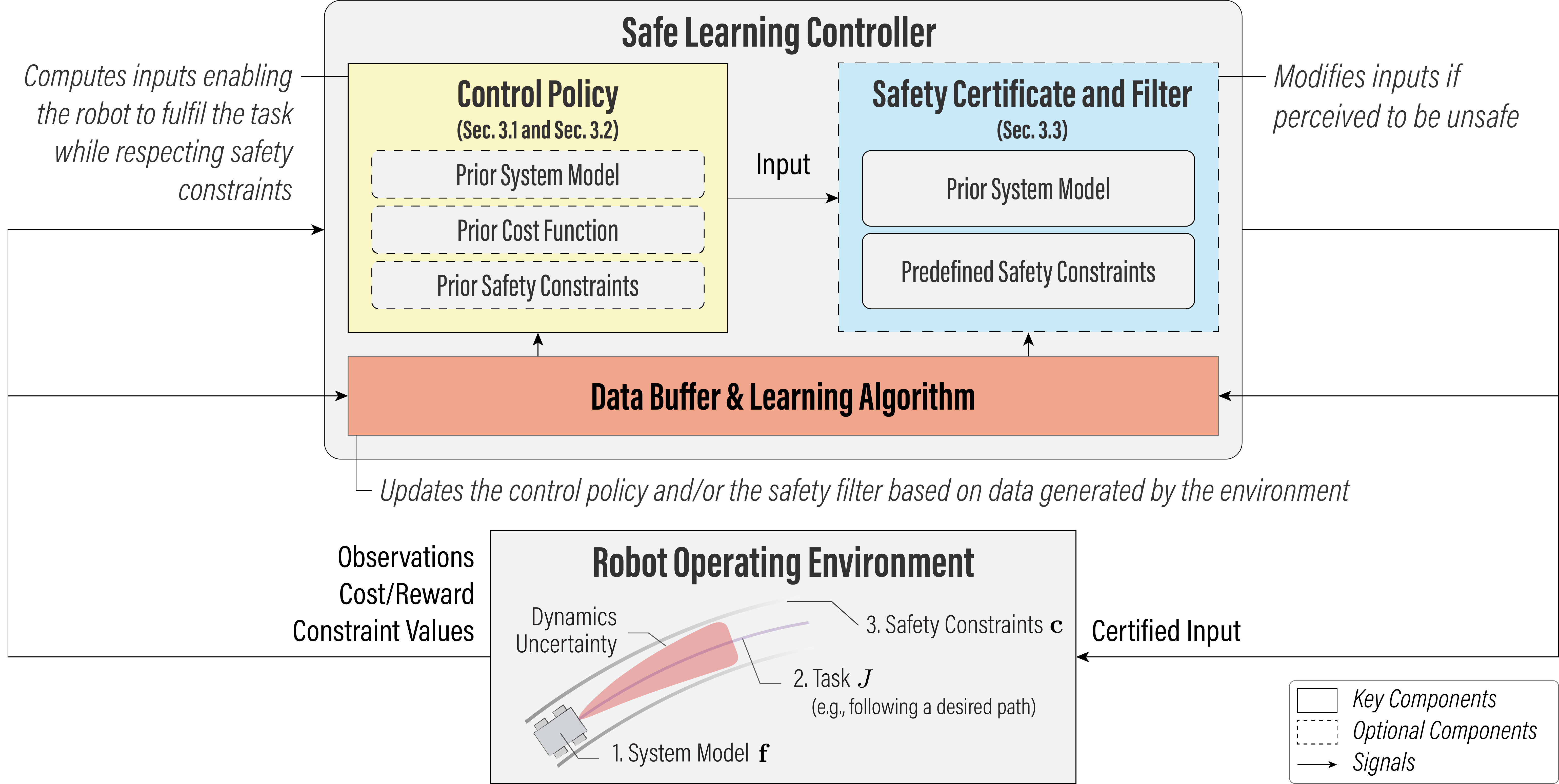}
    \caption{The safe learning control problem is defined by the cost function $J$, the system model $\vec{f}$, and the constraints $\vec{c}$, which may all be initially unknown. Data is used to update the control policy  (see \autoref{sec:uncertain-dyn} and \autoref{sec:methods-rl}) or the safety filter (see~\autoref{sec:certification}).}
    \label{fig:block_diagram}
\end{figure}

We formulate the safe learning control problem as an optimization problem  to capture the efforts of both the control and RL  communities. The optimization problem has three main components (see \textbf{\autoref{fig:block_diagram}}):
\textit{(i)}~a system model that describes the dynamic behavior of the robot, \textit{(ii)}~a cost function that defines the control objective or task goal, and \textit{(iii)}~a set of constraints that specify the safety requirements. The goal is to find a \textit{controller} or \textit{policy} that computes commands (also called \textit{inputs}) that enable the system to fulfil the task while respecting given safety constraints. In general, any of the components could be initially unknown or only partially known. Below, we first introduce each of the three components and conclude by stating the overall safe learning control problem.

\begin{marginnote}[]
\refstepcounter{margin}
\entry{M\arabic{margin}. Transition Probability Function}{The dynamics model in~\autoref{eq:system_model} can be equivalently represented as a state transition probability function~$T_k(\sysstate_{k+1} \,|\, \sysstate_k, \sysinput_k)$, commonly seen in RL approaches~\citep{Sutton2018}.}

\label{mrgn:prob_models}
\end{marginnote}

\subsubsection{System Model} We consider a robot whose dynamics can be represented by the following discrete-time model:
\begin{equation}
    \label{eq:system_model}
\sysstate_{k + 1} = \vec{f}_k(\sysstate_{k}, \sysinput_{k}, \vec{w}_{k}),
\end{equation}
where $k\in\mathbb{Z}_{\ge 0}$ is the discrete-time index, $\sysstate_k\in\set{X}$ is the state with $\set{X}$ denoting the state space, 
$\sysinput_k\in\set{U}$ is the input with $\set{U}$ denoting the input space, 
 $\vec{f}_k$ is the dynamics model of the robot, $\wnoise_k\in\set{W}$ is the process noise distributed according to a distribution~$\mathcal{W}$ (see also~\marginref{mrgn:prob_models}). Throughout this review, we assume direct access to (possibly noisy) measurements of the robot state  $\sysstate_k$ and neglect the problem of state estimation. \autoref{eq:system_model} represents many common robot platforms (e.g., quadrotors, manipulators, and ground vehicles). More complex models (e.g., partial differential equations) may be necessary for other robot designs.

\subsubsection{Cost Function} The robot's task is defined by a cost function. We consider a finite-horizon optimal control problem with time horizon~$N$. Given an initial state $\sysstate_0$, the cost is computed based on the sequence of states $\sysstate_{0:N} =\{\sysstate_{0}, \sysstate_{1}, ..., \sysstate_{N}\}$ and the sequence of inputs $\sysinput_{0:N-1}=\{\sysinput_{0}, \sysinput_{1}, ..., \sysinput_{N-1}\}$:
\begin{equation}
    \label{eq:cost_function}
    J(\sysstate_{0:N}, \sysinput_{0:N-1}) =  l_N (\sysstate_{N}) + \sum_{k=0}^{N-1} l_k(\sysstate_{k},\sysinput_{k}),
\end{equation}
where $l_k: \mathbb{X} \times \mathbb{U} \mapsto \mathbb{R}$ is the stage cost incurred at each time step~$k$, and $l_N: \mathbb{X}\mapsto \mathbb{R}$ is the terminal cost incurred at the end of the $N$-step horizon (see also~\marginref{mrgn:disc_rewards}).\begin{marginnote}[]
\refstepcounter{margin}
\entry{M\arabic{margin}. Discounted Rewards}{The stage cost can be related to the discounted rewards in RL: $l_k = -\gamma^k r_k(\sysstate_k, \sysinput_k)$, where
$r_k: \mathbb{X} \times \mathbb{U} \mapsto \mathbb{R}$ is the reward function, and $\gamma\in [0,1]$ is the discount factor~\citep{Sutton2018}. 
We formulate a cost minimization problem for safe learning control, while RL typically solves a reward maximization problem.
}
\label{mrgn:disc_rewards}
\end{marginnote}The stage and terminal cost functions map the state and input sequences, which may be random variables, to a real number, and may, for example, include the expected value or variance operators.

\subsubsection{Safety Constraints} \label{sec:constraints}

Safety constraints ensure, or encourage, the safe operation of the robot and include: $\emph{(i)}$~\textit{state constraints} $\stateconstraints \subseteq \set{X}$, 
which define the set of safe operating states (e.g., the lane in self-driving), $\emph{(ii)}$~\textit{input constraints} $\inputconstraints \subseteq \set{U}$~(e.g., actuation limits), and $\emph{(iii)}$~\textit{stability guarantees} (e.g., the robot's motion converging to a desired path, see~\marginref{mrgn:stability}).

\begin{marginnote}[]
\refstepcounter{margin}
\entry{M\arabic{margin}. Stability}{There are different notions of stability in the control literature~\cite{khalil2002}. 
Stability generally implies boundedness of the system state. One common notion of stability is  Lyapunov stability, which requires the system to either stay close to (\textit{stable}) or converge to (\textit{asymptotically stable}) a desired state.
}
\label{mrgn:stability}
\end{marginnote}

\begin{figure}\centering
    \includegraphics[draft=false, width=\textwidth]{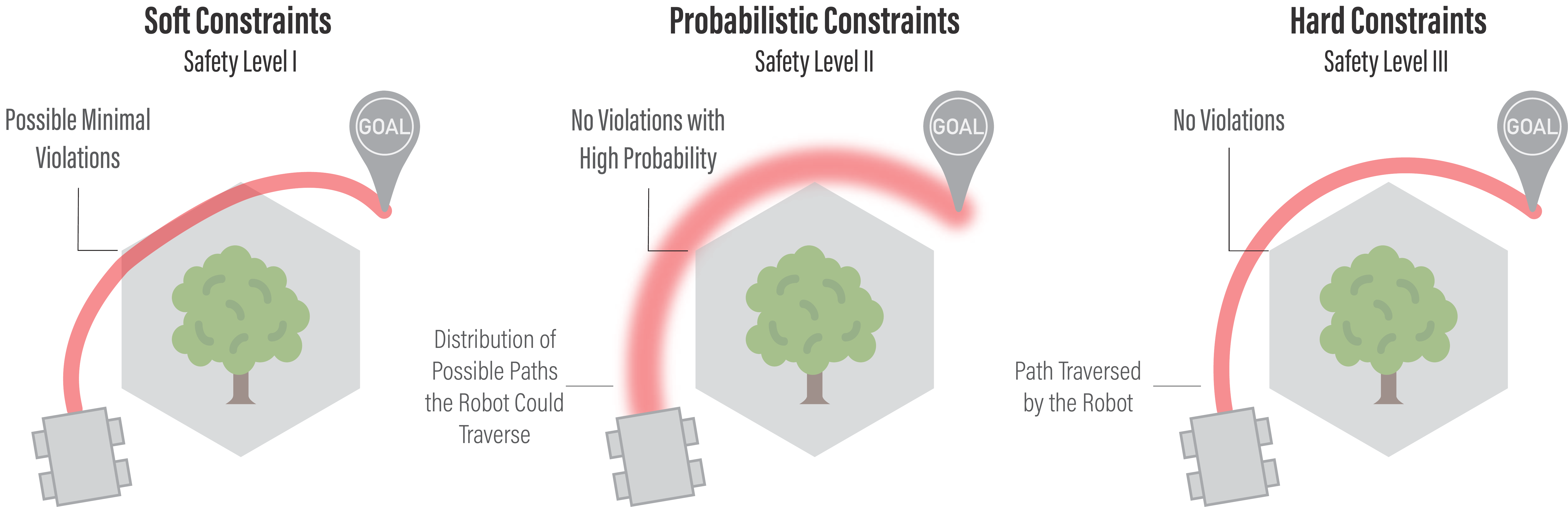}
    \caption{Illustration of the different safety levels.}
    \label{fig:safety_levels}
\end{figure}

To encode the safety constraints, we define ${n_c}$ constraint functions: $\cstr_k(\sysstate_{k}, \sysinput_{k}, \wnoise_k) \in \R^{n_c}$ with each constraint $\constraintsingle_k^j$ being a real-valued, time-varying function.
Starting with the strongest guarantee, we introduce three levels of safety: hard, probabilistic, and soft constraints (illustrated in \textbf{\autoref{fig:safety_levels}}). 
In practice,  safety levels  are often mixed. For example, input constraints are typically hard constraints but state constraints may be soft constraints.

\paragraph*{Safety Level III: Constraint Satisfaction Guaranteed.} 
The system satisfies hard constraints:
\begin{equation}
    \label{eq:hard_constraints}
    \constraintsingle^j_k(\sysstate_{k}, \sysinput_{k}, \wnoise_k) \leq 0\,, 
\end{equation}
for all times $k \in \{0, ..., N\}$ and constraint indexes $j \in \{1, ..., n_c\}$.

\paragraph*{Safety Level II: Constraint Satisfaction with Probability $p$.} 
The system satisfies probabilistic constraints:
\begin{equation}
    \label{eq:chance_constraints}
    \Pr \left(\constraintsingle_k^j\left(\sysstate_{k}, \sysinput_{k}, \wnoise_k\right) \leq 0\right) \geq p^j\,, 
\end{equation}
where $\Pr(\cdot)$ denotes the probability and  $p^j \in (0, 1)$ defines the likelihood of the $j$-th constraint being satisfied, with  $j \in \{1, ..., n_c\}$ and  for all times $k \in \{0, ..., N\}$. The chance constraint in \autoref{eq:chance_constraints} is identical to the hard constraint in \autoref{eq:hard_constraints} for $p^j = 1.$
\paragraph*{Safety Level I: Constraint Satisfaction Encouraged.} 
The system encourages constraint satisfaction. This can be achieved in different ways. One way is to  add a penalty term to the objective function that discourages the violation of constraints with a high cost.
A  non-negative $\epsilon_j$ is added to the right-hand side of the inequality~\autoref{eq:hard_constraints},  for all times $k \in \{0, ..., N\}$ and $j \in \{1, ..., n_c\},$
\begin{align}
    \constraintsingle_k^j(\sysstate_{k}, \sysinput_{k}, \wnoise_k) &\leq \epsilon_j \,, \label{eq:soft_constraints} \end{align}
and an appropriate penalty term $\costsoft (\boldsymbol{\epsilon}) \geq 0$ with $\costsoft(\boldsymbol{\epsilon}) = 0 \iff \boldsymbol{\epsilon} = \vec{0}$ is added to the objective function. The vector $\boldsymbol{\epsilon}$ includes all elements $\epsilon_j$ and is an additional variable of the optimization problem. 
Alternatively, although $\constraintsingle_k^j(\sysstate_{k}, \sysinput_{k}, \wnoise_k)$ is a step-wise quantity, some approaches only aim to provide guarantees on its expected value~$E[\cdot]$ on a trajectory level: 
\begin{align}
    J_{\constraintsingle^j} = E\left[\sum_{k=0}^{N-1} \constraintsingle_k^j\left(\sysstate_{k}, \sysinput_{k}, \wnoise_k\right)\right] \leq d_j \,, \label{eq:expected_constraints}
\end{align}
where $J_{\constraintsingle^j}$ represents the expected total constraint cost, and $d_j$ defines the corresponding constraint threshold. The constraint function can optionally be discounted as $\gamma^k \constraintsingle_k^j\left(\sysstate_{k}, \sysinput_{k}, \wnoise_k\right)$, similar to the stage cost (see \marginref{mrgn:disc_rewards}). 

\subsubsection{Formulation of the Safe Learning Control Problem}
\label{subsubsec:problem_formulation}
The functions introduced above, i.e., the system model $\dyn$, the constraints  $\cstr$, and the cost function $\cost$,  represent the \textit{true} functions of the robot control problem. In practice, $\dyn$,  $\cstr$, and $\cost$ may be unknown or partially known. 
 Without loss of generality, we assume that each of the true functions $\dyn$,  $\cstr$, and $\cost$ can be decomposed into a nominal component $\bar{(\cdot)}$, reflecting our prior knowledge, and an unknown component $\hat{(\cdot)}$, to be learned from data. For instance, the dynamics model~$\dyn$ can be  decomposed as 
 \begin{equation}
     \dyn_k(\sysstate_{k}, \sysinput_{k}, \vec{w}_{k}) = \dynprior_k(\sysstate_{k}, \sysinput_{k}) + \dynlearned_k(\sysstate_{k}, \sysinput_{k}, \vec{w}_{k}),
 \end{equation}
 where $\dynprior$ is the prior dynamics model and $\dynlearned$ are the uncertain dynamics.

\textit{Safe learning control} (SLC) leverages our prior knowledge $\prior =\{\dynprior,  \cstrprior, \costprior\}$  and the data collected from the system $\data = \{\sysstate^{(i)}, \sysinput^{(i)},  \cstr^{(i)}, \coststep^{(i)}\}_{i=0}^{i=D}$ to find a policy (or controller)~$\policy_k(\sysstate_{k})$ that achieves the given task while respecting all safety constraints,
\begin{equation}
\label{eqn:slc_definition}
    \text{SLC}: (\mathcal{P}, \mathcal{D}) \mapsto \policy_k,
\end{equation}
where $(\cdot)^{(i)}$ denotes a sample of a quantity $(\cdot)_k$ and $D$ is the data set size. 
 More specifically, we aim to find a policy $\policy_k$ that best approximates the true optimal policy $\policy^{\star}_k$, which is the solution to the following optimization problem:
\begin{subequations}
    \label{eq:problem_statement}
    \begin{align}
      \cost^{\policy^\star}(\bar{\sysstate}_0) = \min_{\policy_{0:N-1}, \boldsymbol{\epsilon}} & \quad \cost(\sysstate_{0:N}, \sysinput_{0:N-1}) + \costsoft(\boldsymbol{\epsilon})\label{eq:objective_ps}\\
\text{s.t.} &  \quad \sysstate_{k + 1} = \dyn_k(\sysstate_{k}, \sysinput_{k}, \vec{w}_{k}),\, \quad\vec{w}_{k} \sim \dist{W},\, \quad \forall k \in \{0, ..., N - 1\},\label{eq:dynamics_ps}\\
& \quad \sysinput_{k} = \policy_k(\sysstate_{k}), \label{eq:policy_ps} \\
     &\quad \sysstate_0 = \bar{\sysstate}_0,\, \label{eq:initial_state}\\
& \quad \text{Safety constraints according to either}\label{eq:state_input_constraints_ps}\\
      &\hspace{1em}\text{\autoref{eq:hard_constraints}, \autoref{eq:chance_constraints}, \autoref{eq:soft_constraints} or \autoref{eq:expected_constraints}, and $\boldsymbol{\epsilon} \geq \vec{0}$, }\nonumber
    \end{align}
\end{subequations}
where $\bar{\sysstate}_0\sim\mathcal{X}_0$ is the initial state with $\mathcal{X}_0$ being the initial state distribution,
and $\boldsymbol{\epsilon}$ and $\costsoft$ are introduced to account for the soft safety constraint case (Safety Level~I, \autoref{eq:soft_constraints}) and are set to zero, for example,  if  only hard and probabilistic safety constraints are considered (Safety Levels II and III).

\subsection{A Control Theory Perspective}
\label{subsec:control_background}

Safe decision making under uncertainty has a long history in the field of control theory. 
Typical assumptions are that a model of the system is available and it is either parameterized by an unknown parameter or it has bounded unknown dynamics and noise. 
While the control approaches are commonly formulated using continuous-time dynamic models, they are usually implemented in discrete time with sampled inputs and measurements~(see~\marginref{mrgn:cont_time}). 

\begin{marginnote}[]
\refstepcounter{margin}
\entry{M\arabic{margin}. Continuous Time}{
Many classical control approaches rely on a continuous-time formulation. Consider the uncertain time-varying, continuous-time model $\dot{\vec{x}} = \vec{f}_{t}(\vec{x}, \vec{u}, \vec{w})$, where $\vec{x}, \vec{u}$ and $\vec{w}$ are functions of time~$t$. We can recover our discrete state through sampling $\vec{x}_k = \vec{x}(k \Delta t)$, where $\Delta t$ is the sampling time. Similarly, the discrete control input is $\vec{u}_k = \vec{u}(k \Delta t)$ and often kept constant~\cite{astorm2011} over the time interval $\left[k \Delta t, (k + 1) \Delta t \right)$.}
\label{mrgn:cont_time}
\end{marginnote}

\textbf{Adaptive control} considers systems with parametric uncertainties (see  \marginref{mrgn:param_uncertainty}) and adapts the controller parameters  online  to  optimize  performance.
Adaptive control requires knowledge of the parametric form of the uncertainty~\cite{Sastry2011} and typically considers a dynamics model 
that is affine in~$\sysinput$ and the uncertain parameters 
$\params \in \Theta$:
\begin{equation}
    \vec{x}_{k + 1} = \dynprior_{\sysstate}(\sysstate_k) + \dynprior_{\sysinput}(\sysstate_k) \sysinput_k + \dynprior_{\params}(\sysstate_k)\params \,,
    \label{eq:adaptive_control_structure}
\end{equation}
where $\dynprior_{\sysstate}, \dynprior_{\sysinput}$, and $\dynprior_{\params}$ are known functions often derived from first principles and $\Theta$ is a possibly bounded parameter set. The control input is
$\sysinput_k = \policy(\sysstate_k, \hat{\params}_k)$, which is parameterized by~$\hat{\params}_k$.
The parameter~$\hat{\params}_k$ is adapted by using either a \textit{Lyapunov function} to guarantee that the closed-loop system is stable (see \marginref{mrgn:lyapunov}) or  \textit{Model Reference Adaptive Control~(MRAC)} to make the system behave as a predefined stable reference model~\cite{Sastry2011}. 
Adaptive control is typically limited to  parametric uncertainties and relies on a specific model structure.
Moreover, adaptive control approaches tend to ``overfit'' to the latest observations and convergence to the true parameters is generally not guaranteed~\citep{Sastry2011,nguyen2011model}.
These limitations motivate the learning-based adaptive control approaches in \autoref{sec:learning_adaptation}.

\begin{marginnote}[]
\refstepcounter{margin}
\entry{M\arabic{margin}. Parametric Uncertainty}{A \textit{parametric model} depends on a finite number of parameters that  may have  a physical interpretation or reflect  our prior knowledge about the system structure in other ways. \textit{Parametric uncertainty} is the uncertainty in the model parameters.}
\label{mrgn:param_uncertainty}
\end{marginnote}

\label{sec:robust_control}

\textbf{Robust control} is a control design technique that guarantees stability for pre-specified bounded disturbances, which can include unknown dynamics and noise. In contrast to adaptive control, which adapts to the parameters currently present, robust control finds a suitable controller for all possible disturbances and keeps the controller unchanged after the initial design.
Robust control is largely limited to linear, time-invariant~(LTI) systems with linear nominal model $\dynprior(\sysstate_k, \sysinput_k) = \linearpriorA \vec{x}_k + \linearpriorB \vec{u}_k$ and unknown dynamics $\dynlearned_k(\sysstate_k, \sysinput_k, \wnoise_k)=\linearunknownA \vec{x}_k + \linearunknownB \sysinput_k + \wnoise_k   \in \set{D}$,
with $\set{D}$ being known and bounded,  and $\linearpriorA$, 
$\linearpriorB$,
$\linearunknownA$,
$\linearunknownB$ being static matrices of appropriate size. 
That is,   
\begin{equation}
            \sysstate_{k + 1} = (\linearpriorA + \linearunknownA) \sysstate_k + (\linearpriorB + \linearunknownB) \sysinput_k + \wnoise_k\,. 
            \label{eq:linear_robust}
\end{equation}
Robust control design techniques, such as robust $\Hfunc_\infty$- and $\Hfunc_2$-control design~\cite{Zhou1996}, yield controllers that are robustly stable for all $\dynlearned_k \in \set{D}$. 
Robust control can be extended to nonlinear systems whose dynamics can be decomposed into a linear nominal model~$\dynprior$ and a nonlinear function~$\dynlearned$ with known bound~$\dynlearned \in \set{D}$~\cite{dullerud2005}. 

\label{sec:robust_mpc}
\textbf{Robust Model Predictive Control}~(MPC) extends classical adaptive and robust control by additionally guaranteeing state and input constraints, $\sysstate_{k}\in\stateconstraints$ and $\sysinput_{k}\in\inputconstraints$, for all possible bounded disturbances~$\dynlearned \in \set{D}$. 
At every time step $k$, MPC solves a constrained optimization problem over a control input sequence $\sysinput_{k: k + H-1}$ for a finite horizon $H$, and applies the first optimal control input to the system and then re-solves the optimization problem in the next time step based on the current state~\cite{Rawlings2017}.\begin{marginnote}[]
\refstepcounter{margin}
\entry{M\arabic{margin}. Lyapunov Function}{
Lyapunov functions are used to analyze the stability of a dynamical system~\cite{khalil2002}.
Consider  a  closed-loop system under some policy~$\policy(\sysstate)$:
$\sysstate_{k+1} = \dyn_{\policy}(\sysstate_k) = \dyn(\sysstate_k,\policy(\sysstate_k))$ with $\dyn_{\policy}(\sysstate_k)$ being Lipschitz continuous.
A Lipschitz continuous positive definite function $\lyapunov : \set{X} \to \R_{\geq 0}$ with $\lyapunov(\vec{0}) = 0$ and $\lyapunov(\sysstate) > 0, \forall \sysstate \neq \vec{0}$, is a Lyapunov function, if $\lyapunov$ maps states under closed-loop state feedback 
to strictly smaller values (i.e., $\Delta \lyapunov(\sysstate)=\lyapunov(\dyn_{\policy}(\sysstate))-\lyapunov(\sysstate)<0$).
This implies that the state converges to the equilibrium at the origin. 
There exists an analogous formulation for continuous-time systems based on the derivatives of $\lyapunov$~\cite{khalil2002}. 
}
\label{mrgn:lyapunov}
\end{marginnote}A common approach in robust MPC (RMPC) is \textit{tube-based MPC}~\cite{Mayne2005}, which uses a nominal prediction model $\dynprior(\sysstate_k, \sysinput_k)$ in the MPC optimization and tightens the constraints to account for unmodeled dynamics. A stabilizing controller keeps the true state inside a bounded set of states around the nominal state, called $\emph{tube}$, for all possible disturbances. 
Since the nominal states satisfy the tightened constraints and the true states stay inside the tube around the nominal states, constraint satisfaction for the true states is guaranteed. 
Tube-based MPC typically considers a linear nominal model~$\dynprior(\bar{\sysstate}_k, \bar{\sysinput}_k) = \linearpriorA \bar{\sysstate}_k + \linearpriorB \bar{\sysinput}_k$ with nominal state~$\bar{\sysstate}_k$ and input~$\bar{\sysinput}_k$.
In its simplest implementation, prior knowledge of set~$\set{D}$ is combined with a known stabilizing linear controller, $\sysinput_{k, \mathrm{stab}} = \vec{K} (\sysstate_k - \bar{\sysstate}_k)$ 
with gain $\vec{K}$,
to determine the bounded tube $\Omega_{\mathrm{tube}}$ from the matrices $\linearpriorA, \linearpriorB, \vec{K}$, and the set $\set{D}$. The stabilizing controller $\sysinput_{k, \mathrm{stab}}$ keeps all potential errors within the tube, 
${\sysstate}_{k} - \bar{\sysstate}_{k} \in \Omega_{\mathrm{tube}}$, for all  $k$. For the nominal model, 
tube-based MPC solves the following constrained optimization problem at every time step $k$ to obtain the optimal sequence $\bar{\sysinput}_{0:H-1}^*$:
\begin{subequations}
    \label{eq:robust_mpc_opti}
    \begin{align}
      \cost_{\mathrm{RMPC}}^*(\bar{\sysstate}_k) = \min_{\bar{\sysinput}_{0:H-1}} & \quad l_H (\rmpcsysstate_{H}) + \sum_{i=0}^{H-1} l_i(\rmpcsysstate_{i},\bar{\sysinput}_{i}) \label{eq:objective_rmpc}\\
\text{s.t.} &  \quad \rmpcsysstate_{i + 1} = \linearpriorA \rmpcsysstate_i + \linearpriorB \bar{\sysinput}_i,\,\quad  \forall i \in \{0, ..., H - 1\} \label{eq:dynamics_rmpc}\\
    &\quad \rmpcsysstate_i \in \stateconstraints \ominus \Omega_{\mathrm{tube}},\quad \bar{\sysinput}_i \in \inputconstraints \ominus \vec{K} \Omega_{\mathrm{tube}}, \label{eq:rmpc_state_input_constraints} \\
     &\quad \rmpcsysstate_0 = \bar{\sysstate}_k,\, \quad\rmpcsysstate_H \in \set{X}_{\mathrm{term}} \,, \label{eq:initial_state_rmpc}
    \end{align}
\end{subequations}
where $\rmpcsysstate_i$ is the open-loop nominal state at time step $k + i$ and the state and input constraints are tightened using the bounded tube $\Omega_{\mathrm{tube}}$~(see~\marginref{mrgn:tighten_constraints}).Combined with the stabilizing control input~$\sysinput_{k, \mathrm{stab}}$, the control input~$\sysinput_k = \sysinput_{k, \mathrm{stab}} + \bar{\sysinput}_0^*$ is applied to the system at every time step. 
Stability and satisfaction of the tightened constraints is guaranteed by selecting the terminal cost $l_H$ in \autoref{eq:objective_rmpc}, such that after the prediction horizon $H$ the nominal state $\rmpcsysstate_H$ is within a terminal constraint set $\set{X}_{\mathrm{term}}$ (see \autoref{eq:initial_state_rmpc}) at which point a known linear controller can be safely applied~\cite{Rawlings2017}.

\begin{marginnote}[]
\refstepcounter{margin}
\entry{M\arabic{margin}. Constraint Tightening}{
\label{mrgn:tighten_constraints}
The constraint sets $\stateconstraints$ and $\inputconstraints$ are  tightened by an error tube $\Omega_{\mathrm{tube}}$ using the Pontryagin  difference, 
$\stateconstraints \ominus \Omega_{\mathrm{tube}} = \{ \sysstate \in \set{X} : \sysstate + \boldsymbol{\omega} \in \stateconstraints, \forall \boldsymbol{\omega} \in \Omega_{\mathrm{tube}} \}$ and 
$\inputconstraints \ominus \vec{K} \Omega_{\mathrm{tube}}$, where
$\vec{K}$ maps all elements in $\Omega_{\mathrm{tube}}$ to the control input space~$\set{U}$ with $\vec{K}\Omega_{\mathrm{tube}}$.
}
\end{marginnote}

Both robust control and robust MPC are conservative, as they guarantee stability and---in the case of MPC---state and input constraints for the worst-case scenario. This usually yields poor performance, as described in~\cite{Hewing2020}. For example, a conservative uncertainty set $\set{D}$ generates a large tube $\Omega_{\mathrm{tube}}$ resulting in tight hard constraints that are prioritized over cost optimization.
Learning-based robust control and RMPC improve performance by using data to \textit{(i)} learn a less conservative state- and input-dependent uncertainty set~$\set{D}$ and/or \textit{(ii)} learn the unknown dynamics $\dynlearned$ and, as a result, reduce the remaining model uncertainty, see \autoref{sec:learning_robust_control} and \autoref{sec:learning_mpc} respectively.

\label{sec:adaptive_control}

 \subsection{A Reinforcement Learning Perspective}
\label{subsec:reinforcement_learning_background}

Reinforcement learning (RL) is the standard machine learning framework to address the problem of sequential decision making under uncertainty. 
Unlike traditional control, RL generally does not rely on an \emph{a priori} dynamics model $\dynprior$ and can be directly applied to uncertain dynamics $\dyn$.
However, 
the lack of explicit assumptions and constraints in many of the works limit their applicability to safe control.
RL algorithms attempt to find $\policy^*$ while gathering data and knowledge of $\dyn$ from interaction with the system---taking random actions, initially, and improving afterwards. 
A long-standing challenge of RL, which hampers safety during the learning stages, is the exploration-exploitation dilemma---that is, whether to \emph{(i)} act greedily with the available data or to \emph{(ii)} explore, which means taking sub-optimal---possibly unsafe---actions $\sysinput$ to learn a more accurate $\dynlearned$.

RL typically assumes that the underlying control problem is a Markov decision process~(MDP). 
An MDP comprises a state space~$\set{X}$, an input (\emph{action}) space~$\set{U}$,
stochastic dynamics (also called \emph{transition model}, see \marginref{mrgn:prob_models}),
and a per-step reward function (see \marginref{mrgn:disc_rewards}). When all the components of an MDP are known (in particular, $\dyn$ and $\cost$ in \autoref{subsec:problem_statement}),
then it solves the problem in Equations~\ref{eq:objective_ps}, \ref{eq:dynamics_ps}, \ref{eq:policy_ps} without the constraints. Dynamic programming~(DP) algorithms such as value and policy iteration can be used to find an optimal policy $\policy^{*}$. Many RL approaches, however, make no assumptions on any part of $\dyn$ being known \emph{a priori}.

\begin{marginnote}[]
\refstepcounter{margin}
\entry{M\arabic{margin}. Value and Action-Value Functions}{
The value function of state $\sysstate_k$ under policy $\policy_k$, $V^{\policy_k}(\sysstate_k$), is the expected return $\cost$ of applying $\policy_k$ from $\sysstate_k$.
Similarly, the action-value function of action $\sysinput_k$ in state $\sysstate_k$ under $\policy_k$, $Q^{\policy_k}(\sysstate_k,\sysinput_k)$, is the expected return $\cost$ when taking the action $\sysinput_k$ at $\sysstate_k$, and then following $\policy_k$.
\label{mrgn:value_fun}
}
\end{marginnote}

We can distinguish model-based RL approaches, which learn an explicit model $\dynlearned$ of the system dynamics $\dyn$ and use it to optimize a policy, from model-free RL algorithms.
The latter algorithms~\cite{Arulkumaran2017} can be broadly categorized as:
\emph{(i)} value function-based methods, learning a value function (see \marginref{mrgn:value_fun}); 
\emph{(ii)} policy-search and policy-gradient methods, directly trying to find an optimal policy $\policy^*$; 
and \emph{(iii)} actor-critic methods, learning both a value function (\emph{critic}) and the policy (\emph{actor}).
We also note that the convergence of these methods has been shown for simple scenarios but is still a challenge for more complex scenarios or when function approximators are used~\cite{dai2018}.

There are multiple practical hurdles to the deployment of RL algorithms in real-world robotics problems~\cite{dulac2019a}, for example,
\textit{(i)} the continuous, possibly high-dimensional
$\mathbb{X}$ and $\mathbb{U}$ in robotics (often assumed to be finite, discrete sets in RL), \textit{(ii)} the stability and convergence of the learning algorithm~\cite{cheng2019b} (necessary, albeit not sufficient, to produce a stable policy, see~\marginref{mrgn:stability}), \textit{(iii)} learning robust policies from limited samples, \textit{(iv)} the interpretability of the learned policy---especially in deep RL when leveraging neural networks (NNs) (see \marginref{mrgn:neural_nets}) for function approximation~\cite{Arulkumaran2017}---and, importantly, \textit{(v)} providing provable safety guarantees.

\begin{marginnote}[]
\refstepcounter{margin}
\entry{M\arabic{margin}. Neural Network}{
A \textit{neural network} (NN) is a computational model with interconnected layers of neurons (parameterized by weights) that can be used to approximate highly nonlinear functions.
}
\label{mrgn:neural_nets}
\end{marginnote}

The exploration-exploitation dilemma can be mitigated using Bayesian inference. This is achieved by computing posterior distributions over the states $\set{X}$, possible dynamics $\dyn$, or total cost $J$ from past observed data~\cite{ghavamzadeh2015a}.
These posteriors provide explicit knowledge
of the problem's uncertainty and
can be used to guide exploration.
In practice, however, full posterior inference is almost always prohibitively expensive
and concrete implementations must rely on approximations.

To achieve constraint  satisfaction (over states or sequences of states) and robustness to different, possibly noisy dynamics $\dyn$,
constrained MDPs and robust MDPs are
extensions of traditional MDPs that more closely resemble the problem statement in \autoref{subsec:problem_statement}.

\textbf{Constrained MDPs} (CMDPs)~\cite{altman1999constrained} extend simple MDPs with constraints and optimize the problem in
\autoref{eq:problem_statement} when \autoref{eq:state_input_constraints_ps} takes the discounted form of \emph{Safety Level I}'s \autoref{eq:expected_constraints}. We refer to the discounted constraint cost $J_{c^j}$ under policy $\policy$ as $J^{\policy}_{c^j}$.
Traditional approaches to solve CMDPs, such as Linear Programming and Lagrangian methods, often assume discrete state-action spaces
and cannot properly scale to complex tasks such as robot control.
Deep RL promises to mitigate this,
yet, applying it to the constrained problem
still suffers from the computational complexity of the off-policy evaluation (\marginref{mrgn:offpolicy_onpolicy}) of trajectory-level constraints~$J_{c^j}$~\cite{achiam2017constrained}.
In \autoref{sec:methods-rl:cmdps}, we present some 
recent advances in CMDP-based work that feature:
\emph{(i)}~integration of deep learning techniques in CMDPs for more complex control tasks; \emph{(ii)}~provable constraint satisfaction throughout the exploration or learning process; and \emph{(iii)}~constraint transformation for the efficient evaluation of $J_{c^j}$ from data collected off-policy.

\begin{marginnote}[]
\refstepcounter{margin}
\entry{M\arabic{margin}. Off-policy, On-policy, and Offline RL}{Off-policy RL improves the value functions estimates with data collected operating under different policies. On-policy methods only use data generated by the current policy.
Offline RL uses data previously collected by an unknown policy.
}
\label{mrgn:offpolicy_onpolicy}
\end{marginnote}

\textbf{Robust MDPs}~\cite{nilim2005a}, inspired by robust control (see \autoref{subsec:control_background}), extend MDPs such that the
dynamics can include parametric uncertainties or disturbances, and the cost of the worst-case scenario is optimized. This is captured by the min-max optimization problem:
\begin{subequations}
    \label{eq:minimax_rl}
    \begin{align}
    J^{\policy} (\bar{\sysstate}_0) = \min_{\policy_{0:N-1}}  \max_{\dynlearned \in \set{D}} &\quad \cost(\sysstate_{0:N}, \sysinput_{0:N-1}) \\
\text{s.t.} & \quad \text{\autoref{eq:dynamics_ps}, \autoref{eq:policy_ps}, \autoref{eq:initial_state}, }
    \end{align}
\end{subequations}
where $\set{D}$ is a given uncertainty set of $\dynlearned$. 
To keep solutions tractable, practical implementations typically restrict $\set{D}$ to certain classes of models. This can limit the applicability of robust MDPs beyond toy problems. 
Recent work~\cite{pinto2017robust,pan2019a,vinitsky2020robust} applied deep RL to robust decision making, targeting key theoretical and practical hurdles such as \textit{(i)} how to effectively model uncertainty with deep neural networks, and \textit{(ii)} how to efficiently solve the min-max optimization (e.g., via sampling or two-player, game-theoretic formulations). These ideas, including adversarial RL and domain randomization, are presented in \autoref{sec:methods-rl:rrl}.

 \subsection{Bridging Control Theory and RL for Safe Learning Control}
\label{subsec:data_in_safe_learning}

\begin{figure}\centering
    \includegraphics[draft=false, width=\textwidth]{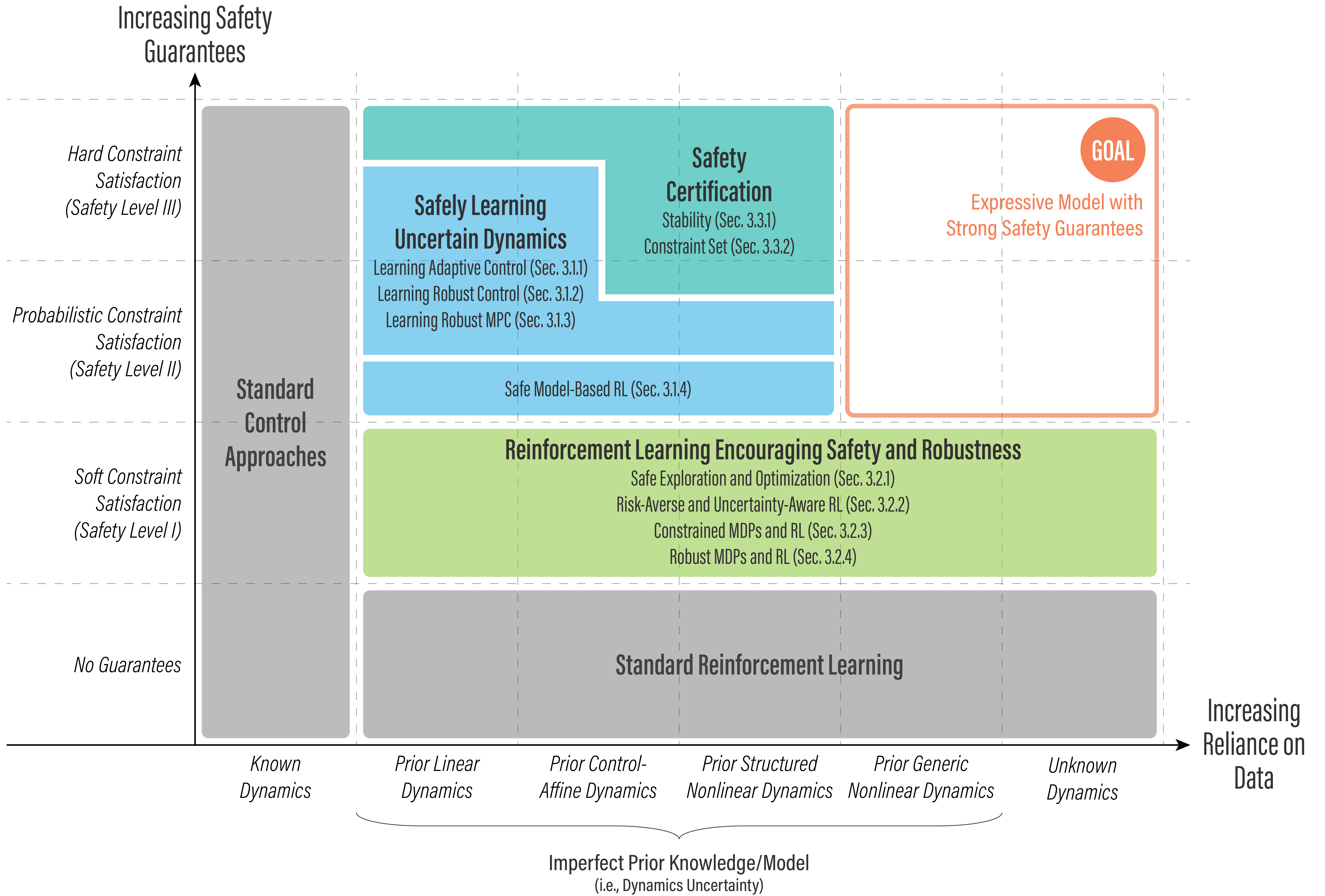}
    \caption{A summary of the safe learning control approaches reviewed in \autoref{sec:methods}.}
    \label{fig:survey_summary}
\end{figure}

When designing a learning-based controller, we typically have two sources of information: \textit{(i)} our prior knowledge, and \textit{(ii)} data generated by the robot system. 
Control approaches  rely on prior knowledge and on assumptions such as  parametric dynamics models to provide safety guarantees. RL approaches typically make use of expressive learning models to extract patterns from data that facilitate learning complex tasks, but these can impede the provision of any formal guarantees. In recent literature, we see an effort from both the control and RL communities to develop safe learning control algorithms, with the  goal of systematically leveraging expressive models for closed-loop control (see \textbf{\autoref{fig:survey_summary}}). Questions that arise from these efforts include: how can control-theoretic tools be applied to expressive machine learning models? Or, how can expressive models be incorporated into  control frameworks?

\begin{marginnote}[]
\refstepcounter{margin}
\entry{M\arabic{margin}. Gaussian Process}{
A \textit{Gaussian process} (GP) is a set of random variables in which any finite number of the random variables can be modeled as a joint Gaussian distribution. Practically, a GP is a probabilistic model specifying a distribution over functions.
}
\label{mrgn:gps}
\end{marginnote}

Expressive learning models  can be categorized as  deterministic  (e.g., standard deep neural networks (DNNs)) or probabilistic  (e.g., Gaussian processes (GPs), see \marginref{mrgn:gps}, and Bayesian linear regression (BLR)). Deep learning techniques such as feed-forward NNs, convolutional NNs, and long short-term memory networks have the advantage of being able to abstract large volumes of data, enabling real-time execution in a control loop. On the other hand, their probabilistic counterparts, such as GPs and BLRs, provide model output uncertainty estimates that can be naturally blended into traditional adaptive and robust control frameworks. We note that there are approaches aiming to combine the advantages of the two types of learning (e.g., Bayesian NNs), and quantifying uncertainty in deep learning is still an active research direction.

  In this review, we focus on approaches that address the problem of safe learning control at two stages: \textit{(i)}  \textit{online adaptation or learning}, where online data is used to adjust the parameters of the controller, the robot dynamics model, the cost function, or the constraint functions during closed-loop operation; and  \textit{(ii)} \textit{offline learning}, where data collected from each trial is recorded and used to update a model in a batch manner in between trials of closed-loop operation. In safe learning control, data is generally used to address the issue of uncertainties in the problem formulation and reduce the conservatism in the system design, while the safety aspect boils down to ``knowing what is not known'' and cautiously accounting for the incomplete knowledge via algorithm design.
 
\section{SAFE LEARNING CONTROL APPROACHES}
\label{sec:methods}

{\color{red}

}

The ability to guarantee safe robot control is inevitably dependent on the amount of prior knowledge available and the type of uncertainties present in the problem of interest. In this section, we discuss approaches for \textit{safe learning control in robotics} based on the following three categories (see \textbf{\autoref{fig:block_diagram}} and \textbf{\autoref{fig:survey_summary}}):

\begin{itemize}
    \item \textbf{Learning uncertain dynamics to safely improve performance:} The first set of works we review relies on an \textit{a priori} model of the robot dynamics. The robot's performance  is improved by learning the uncertain dynamics from data.  Safety  is typically guaranteed based on standard control-theoretic frameworks, achieving Safety Level II or III.
    \item \textbf{Encouraging safety and robustness in RL:} The second set of works encompasses approaches that usually do not have knowledge of  an \textit{a priori} robot model or the safety constraints. Rather than providing hard safety guarantees, these approaches encourage safe robot operation (Safety Level I), for example, by penalizing dangerous actions. \item \textbf{Certifying learning-based control under dynamics uncertainty:} The last set of works aims to provide safety certificates for  learning-based controllers that do not inherently consider safety constraints. 
    These approaches modify the learning controller output by either constraining the control policy,
    leveraging a known safe backup controller,
    or modifying the controller output directly to achieve stability and/or constraint satisfaction. They typically achieve Safety Level II or III.  
\end{itemize}
\textbf{\autoref{fig:survey_summary}} categorizes the approaches reviewed in this section based on their safety level and reliance on data. A more detailed summary of the approaches can be found in~\textbf{\autoref{tab:summary}} of \aref{appendix:appendix}.

\subsection{Learning Uncertain Dynamics to Safely Improve Performance}
\label{sec:uncertain-dyn}
In this section, we consider approaches that improve the robot's performance by learning the uncertain dynamics and provide safety guarantees (Safety Level II or III) via control frameworks such as adaptive control, robust control, and robust MPC (outlined in \autoref{subsec:control_background}). These approaches make assumptions about the unknown parts of the problem (e.g., Lipschitz continuity) and often rely on a known model structure (e.g., control-affine or linear system with bounded uncertainty) to prove stability and/or constraint satisfaction~(see \textbf{\autoref{fig:survey_summary}}).

\subsubsection{Integrating Machine Learning and Adaptive Control}
\label{sec:learning_adaptation}There are three main ideas to incorporating online machine learning into traditional adaptive control (see \autoref{subsec:control_background}), each with its own distinct benefits. These include:  \textit{(i)} using black-box machine learning models to accommodate nonparametric unknown dynamics, \textit{(ii)} using probabilistic learning and explicitly accounting for the learned model uncertainties, in order to achieve cautious adaptation, and \textit{(iii)} augmenting adaptive control with deep learning approaches for experience memorization in order to minimize the need for re-adaptation.

\subheading{Learning Nonparametric Unknown Dynamics with Machine Learning Models.}
 One goal of integrating adaptive control and machine learning is to improve the performance of a robot subject to nonparametric dynamics uncertainties. As opposed to \autoref{eq:adaptive_control_structure}, we consider a system with nominal dynamics~$\dynprior(\sysstate_k, \sysinput_k) = \linearpriorA \sysstate_k + \linearpriorB \sysinput_k$ and unknown dynamics~$\dynlearned(\sysstate_k, \sysinput_k) = \linearpriorB\:\inputdelta_k(\sysstate_k)$, where $\psi_k(\sysstate_k)$ is an unknown nonlinear function  without an obvious parametric structure. 
Learning-based MRAC approaches~\cite{cooper2014use,Gahlawat2020} make the uncertain system behave as the linear nominal model $\dynprior$ by using a combination of $\mathcal{L}_1$ adaptation~\citep{Hovakimyan2010} and online learning to approximate $\inputdelta_k(\vec{x}_k)$ by $\inputdeltalearned_k(\vec{x}_k) = \inputdeltalearned_{\mathcal{L}_1,k} + \inputdeltalearned_{\text{learn},k}(\vec{x}_k)$, where $\inputdeltalearned_{\mathcal{L}_1,k}$ is the input computed by the $\mathcal{L}_1$ adaptive controller and $\inputdeltalearned_{\text{learn},k}$ is the input computed by the learning module, which can be an NN~\cite{cooper2014use} or a GP~\cite{Gahlawat2020}.
The estimated $\inputdeltalearned_k(\vec{x}_k)$ is then used in the controller~$\policy_k(\sysstate_k)$ to account for the unknown nonlinear dynamics $\inputdelta_k(\vec{x}_k)$, improving a linear nominal control policy designed based on~$\dynprior$. 
In these approaches, fast adaptation and stability guarantees are provided by the $\mathcal{L}_1$ adaptation framework (Safety Level~III), while the learning module provides additional flexibility to capture the unknown dynamics. As shown in~\cite{cooper2014use}, the addition of learning improves the performance of the standard $\mathcal{L}_1$ adaptive controller and allows fast adaptation to be achieved at a lower sampling rate.

\subheading{Cautious Adaptation with Probabilistic Model Learning.} 
Another stream of adaptive control approaches leverage probabilistic models to achieve cautious adaptation by weighting the contribution of the learned model based on the model output uncertainty. We consider a system with nominal dynamics $\dynprior(\sysstate_k,\sysinput_k)=\dynprior_{\sysstate}(\sysstate_k) + \dynprior_{\sysinput}(\sysstate_k)\sysinput_k$ and unknown dynamics of the same form $\dynlearned(\sysstate_k,\sysinput_k)=\dynlearned_{\sysstate}(\sysstate_k) + \dynlearned_{\sysinput}(\sysstate_k)\sysinput_k$, where $\dynlearned_{\sysstate}(\sysstate_k)$ and $\dynlearned_{\sysinput}(\sysstate_k)$ are unknown nonparametric nonlinear functions. In a model inversion-based MRAC framework, an approximate feedback linearization is achieved via the nominal model to facilitate the design of MRAC, and a GP-based adaptation approach is used to compensate for feedback linearization errors due to the unknown dynamics~\cite{grande2014experimental}.
To account for the uncertainty in the GP model learning, 
the controller relies on the GP model
only if the confidence in the latter is high: $\policy_k(\sysstate_k) = \policy_\text{nom}(\sysstate_k) - \gamma(\sysstate_k,\sysinput_k)\:\policy_{\text{learn},k}(\sysstate_k)$, where $\policy_\text{nom}(\sysstate_k)$ is the control policy designed based the nominal model, $\policy_{\text{learn},k}(\sysstate_k)$ is the adaptive component designed based on the GP model, and $\gamma(\sysstate_k,\sysinput_k)\in[0,1]$ is a scaling factor with 0 indicating low confidence in the GP.
The stability of the overall system (Safety Level~III) is guaranteed via a stochastic stability analysis, and the efficacy of the approach has been demonstrated in quadrotor experiments~\citep{chowdhary2014bayesian,grande2014experimental}.

\subheading{Memorizing Experience with Deep Architectures.}
Apart from compensating for nonlinear and nonparametric dynamics uncertainties, deep learning approaches have also been applied to adaptive control for ``memorizing'' generalizable feature functions as the system adapts.
In particular, an asynchronous DNN adaptation approach is proposed in \citep{joshi2019deep, joshi2020asynchronous}. Similar to~\cite{cooper2014use,Gahlawat2020}, we consider a linear nominal model~$\dynprior(\sysstate_k, \sysinput_k) = \linearpriorA \sysstate_k + \linearpriorB \sysinput_k$ and  unknown nonlinear dynamics $\dynlearned(\sysstate_k, \sysinput_k) = \linearpriorB\:\inputdelta_k(\sysstate_k)$ with $\inputdelta_k(\sysstate_k)$ being an unknown nonlinear function. In the proposed approach, the last layer of the DNN is updated at a higher frequency for fast adaptation, while the inner layers are updated at a lower frequency to ``memorize'' pertinent features for the particular operation regimes. 
To provide safety guarantees, the authors derive an upper bound on the sampling complexity of the DNN to achieve a prescribed level of modelling error and leverage this result to show that Lyapunov stability of the adapted system can be guaranteed (Safety Level~III) by ensuring the modelling error of the DNN is lower than a given bound.
In contrast to other MRAC approaches, which do not usually retain a memory of the past experience, the inner layers of the asynchronous DNN in \citep{joshi2019deep, joshi2020asynchronous}  store relevant features that facilitate adaptation when similar scenarios arise in the future.

\subsubsection{Learning-based Robust Control}
\label{sec:learning_robust_control}
Learning-based robust control improves the performance of classical robust control in \autoref{subsec:control_background} by using data to improve the linear dynamics model and reduce the uncertainty in \autoref{eq:linear_robust}.

\subheading{Using a Gaussian Process Dynamics Model for Linear Robust Control.} The conservative performance of robust control in \autoref{subsec:control_background} is improved by updating the linear dynamics model and uncertainty in \autoref{eq:linear_robust} with a Gaussian Process \cite{Berkenkamp2015a} (GP). The unknown nonlinear dynamics $\dynlearned(\sysstate_k, \sysinput_k)$ are learnt as a GP, which is then linearized about an operating point. By linearizing the GP (as opposed to directly fitting a linear model) data close to the operating point are prioritized. The uncertain linear dynamics in \autoref{eq:linear_robust} are assumed to be modelled as $\linearunknownA = \vec{A}_0 +  \tilde{\vec{A}} \mathbf{\Delta}$, $\linearunknownB = \vec{B}_0 + \tilde{\vec{B}} \mathbf{\Delta}$, where  $\vec{A}_0$ and $\vec{B}_0$ are obtained from the linearized GP mean, $\tilde{\vec{A}}$ and $\tilde{\vec{B}}$ are obtained from the linearized GP variance (often two standard deviations), and $\mathbf{\Delta}$ represents a matrix with elements taking any value in the range of $[-1, +1]$. Further performance improvement is achieved by modelling $\tilde{\vec{A}}$ and $\tilde{\vec{B}}$ as state dependent \cite{holicki2021a}. Additionally, \cite{vonrohr2021probabilistic} was able to achieve better performance than \cite{Berkenkamp2015a} by leveraging the GP's distribution, while maintaining the same level of safety. The main advantage of these approaches is that they can robustly guarantee Safety Level II stability while improving performance. This is achieved by shrinking the GP uncertainty as more data is added, which improves the linear model $\vec{A}_0$ and $\vec{B}_0$ and reduces the uncertain component $\tilde{\vec{A}}$ and $\tilde{\vec{B}}$. This has been shown on a quadrotor~\cite{Berkenkamp2015a}. However, these approaches are limited to stabilization tasks and do not account for state and input constraints.

\subheading{Exploiting Feedback Linearization for Robust Learning-Based Tracking.} Trajectory tracking convergence, as opposed to the simpler stabilization task performed in \cite{Berkenkamp2015a}, is guaranteed by exploiting the special structure of exactly feedback linearizable systems \cite{Helwa2019}. This structure assumes that the nonlinear system dynamics in \autoref{eq:system_model} can be described by a linear nominal model, where $\linearpriorA$ and $\linearpriorB$ have an integrator chain structure. It also assumes that the unknown dynamics are $\dynlearned(\sysstate_k, \sysinput_k) = \linearpriorB \psi(\sysstate_k, \sysinput_k)$, where $\psi(\sysstate_k, \sysinput_k)$ is an unknown invertible function. A probabilistic upper bound is obtained for $\psi(\sysstate_k, \sysinput_k)$ by learning this function as a GP. A robust linear controller is designed for the uncertain system based on this learnt probabilistic bound. Further performance improvement is achieved by also updating the feedback linearization \cite{greeff2021a} through improving the estimate of the inverse $ \psi^{-1}(\cdot)$. These approaches have been applied to trajectory tracking, with Safety Level II,  on Lagrangian mobile manipulators in \cite{Helwa2019} and quadrotor models in \cite{greeff2021a}. However, they hinge on this special structure and cannot account for state and input constraints.

\subsubsection{Reducing Conservatism in Robust MPC with Learning and Adaptation}
\label{sec:learning_mpc}
The conservative nature of robust MPC (\autoref{subsec:control_background}) is improved---while still satisfying input and state constraints---through \textit{(i)} robust adaptive MPC, which adapts to parametric uncertainty, and \textit{(ii)} learning-based robust MPC, which learns the unknown dynamics~$\dynlearned$ or, in one case, cost~$\costlearned$.

\subheading{Robust Adaptive MPC.}
Robust adaptive MPC assumes parametric uncertainties, and \textit{(i)} uses data to reduce the set of possible parameters over time or \textit{(ii)} uses an inner-loop adaptive controller and an outer-loop robust MPC. This leads to improved performance compared to standard robust MPC (see \autoref{subsec:control_background}), while satisfying hard constraints~(Safety Level III).

The first set of approaches considers stabilization tasks, where the full system dynamics, \autoref{eq:adaptive_control_structure}, are assumed to be linear and stable~\cite{Tanaskovic2014} or linear and unstable~\cite{Lorenzen2019} with uncertain parameter~$\params \in \boldsymbol{\Theta}_0$:
\begin{equation}
\vec{x}_{k + 1} = (\linearpriorA + \linearunknownA(\params)) \sysstate_k + (\linearpriorB + \linearunknownB(\params)) \sysinput_k + \wnoise_k \,.
    \label{eq:adaptive_mpc_sys}
\end{equation}
The process noise and the parameters are assumed to be bounded by known sets $\set{W}$ and the initially conservative $\boldsymbol{\Theta}_0$, respectively. Given $\set{W}$ and $\boldsymbol{\Theta}_0$, we can derive a conservative upper bound on the uncertain dynamics $\dynlearned(\sysstate_k, \sysinput_k, \wnoise_k, \params) = \linearunknownA(\params)\sysstate_k + \linearunknownB(\params) \sysinput_k + \wnoise_k \in \set{D}_0$, where $\set{D}_0$ is a compact set determined from $\set{W}$ and $\boldsymbol{\Theta}_0$. To guarantee constraint satisfaction, tube-based MPC (see \autoref{subsec:control_background}) is applied, where the initial tube~$\Omega_{\mathrm{tube,0}}$ is based on~$\set{D}_0$. To reduce the conservatism of the approach, an adaptive control method is introduced to improve the estimate of the parameter set~$\boldsymbol{\Theta}_k$ and  reduce the size of the tube~$\Omega_{\mathrm{tube,k}}$ at each time step~$k$.
This idea has been extended to stochastic process noise for probabilistic constraint satisfaction~(Safety Level II)~\cite{Bujarbaruah2018}, time-varying parameters~\cite{Bujarbaruah2019}, and linearly parameterized uncertain nonlinear systems~\cite{Goncalves2016, Kohler2020}. Further performance improvement is achieved by combining robust adaptive MPC with iterative learning~\cite{Rosolia2018b}, which updates the terminal constraint set~$\set{X}_{\mathrm{term}}$ in \autoref{eq:initial_state_rmpc} and the terminal cost ~$\coststep_H$ in \autoref{eq:objective_rmpc} after every full iteration using the closed-loop state trajectory and cost~\cite{Bujarbaruah2018c}. In the second approach, an underlying MRAC~(see~\autoref{subsec:control_background}) is used to make the closed-loop system dynamics resemble a linear reference model with bounded disturbance set~$\set{D}$~\cite{Pereida2021}. This linear model and its bounds are then used in an outer-loop robust MPC to achieve fast stabilization in the presence of model errors.

\subheading{Learning-Based Robust MPC.} \label{sec:learning_robust_mpc}
Learning-based robust MPC  uses data to \textit{(i)} improve the unknown dynamics estimate, \textit{(ii)} reduce the uncertainty set, or \textit{(iii)} update the cost to avoid states with high uncertainty. Unlike robust adaptive control, learning-based robust MPC considers nonparameterized systems.

Under the assumption of a linear nominal model $\dynprior(\sysstate, \sysinput) = \linearpriorA \sysstate + \linearpriorB \sysinput$ and bounded unknown dynamics $\dynlearned(\sysstate_k, \sysinput_k) \in \set{D}$, the unknown dynamics can be safely learned from data, e.g., using a NN~\cite{Aswani2013}.
Robust constraint satisfaction, \autoref{eq:rmpc_state_input_constraints}, is guaranteed by using tube-based MPC for the linear nominal model and performance improvement is achieved by optimizing over the control inputs for the combined nominal and learned dynamics. 
If the bounded unknown dynamics are assumed to be state-dependent $\dynlearned(\sysstate_k) \in \set{D}(\sysstate_k)$, instead of using a constant tube~$\Omega_{\mathrm{tube}}$, the state constraints in \autoref{eq:rmpc_state_input_constraints} can be tightened based on the state-dependent uncertainty set~$\set{D}(\sysstate)$~\cite{Soloperto2018}. 
In a numerical stabilization task, a GP is used to model $\dynlearned(\sysstate_k)$ and its covariance determines $\set{D}(\sysstate)$. This yields less conservative, probabilistic~state constraints~(Safety Level II).

Learning-based robust MPC can be extended to nonlinear nominal models. Typically, 
a GP is used to learn the unknown dynamics $\dynlearned(\sysstate, \sysinput)$: the mean updates the dynamics model in \autoref{eq:dynamics_rmpc} and the state- and input-dependent uncertainty set~$\set{D}(\sysstate, \sysinput)$ is derived from the GP's covariance and contains the true uncertainty with high probability. Similarly, the state- and input-dependent tube in \autoref{eq:rmpc_state_input_constraints} is determined from the uncertainty set $\set{D}(\sysstate, \sysinput)$.  The main challenge, compared to using a linear nominal model, is the uncertainty propagation over the prediction horizon in the MPC because Gaussian uncertainty (obtained from the GP) is no longer Gaussian when propagated through the nominal nonlinear dynamics. Approximation schemes are required, such as using a sigma-point transform~\cite{Ostafew2016a}, linearization~\cite{Hewing2019}, exact moment matching~\cite{kamthe2018a}, or ellipsoidal uncertainty set propagation~\cite{Koller2019b}. Additionally, further approximations~(e.g., fixing the GP's covariances over the prediction horizon~\cite{Hewing2019}) are usually required to achieve real-time implementation. These approximations can lead to violations of the probabilistic constraints~(Safety Level~II) in \autoref{eq:state_input_constraints_ps}. 
An alternative approach to address the challenges of GPs uses a NN regression model that predicts the quantile bounds of the tail of a state trajectory distribution for a tube-based MPC~\cite{Fan2020}. However, this approach is hindered by typically nonexhaustive training data sets, which can lead to the underestimation of the tube size.

\begin{marginnote}[]
\refstepcounter{margin}
\entry{M\arabic{margin}. Lipschitz Continuity}{
A function $\mathbf{h}: \mathbb{A} \mapsto \mathbb{B}$ is said to be \textit{Lipschitz continuous} on $\mathbb{A}$ if $(\exists \rho > 0)$ such that for any $\mathbf{a}_1,\mathbf{a}_2\in\mathbb{A}$ the condition $||\mathbf{h}(\mathbf{a}_1)-\mathbf{h}(\mathbf{a}_2)||\le\rho||\mathbf{a}_1-\mathbf{a}_2||$ holds, where $\mathbb{A}$ and $\mathbb{B}$ are the domain and codomain of $\mathbf{h}$, and $||\cdot||$ denotes any $l_p$-norm of a vector. The constant $\rho$ is called a \textit{Lipschitz constant} of $\mathbf{h}$.
}
\label{mrgn:lipschitz_cont}
\end{marginnote}

For repetitive tasks, another approach is to adjust the cost function based on data instead of the system model. The predicted cost error is learned from data using the difference between the predicted cost at each time step and the actual closed-loop cost at execution. By adding this additional term to the cost function, the MPC penalizes states that had previously resulted in higher closed-loop cost than expected~\autoref{eq:objective_rmpc}~\cite{mckinnon2020a}, resulting in reliable performance despite model errors.

\begin{marginnote}[]
\refstepcounter{margin}
\entry{M\arabic{margin}. Region of Attraction}{
Let $\sysstate^*$ be an equilibrium of a  closed-loop system under some policy~$\policy(\sysstate)$:
$\sysstate_{k+1} = \dyn_{\policy}(\sysstate_k) = \dyn(\sysstate_k,\policy(\sysstate_k))$ with $\dyn_{\policy}(\sysstate_k)$ being Lipschitz continuous. The \textit{region of attraction} (ROA) of $\sysstate^*$ is the set of states from which the system will converge to $\sysstate^*$.
}
\label{mrgn:roa}
\end{marginnote}

\subsubsection{Safe Model-Based RL with \emph{A Priori} Dynamics}
\label{sec:smbrl}
Safe model-based RL augments model-based RL~(see~\autoref{subsec:reinforcement_learning_background})  with safety guarantees~(see \textbf{\autoref{fig:survey_summary}}).  
Stability can be probabilistically guaranteed ~(Safety Level~II) under the assumption that the known nominal model~$\dynprior(\sysstate_k, \sysinput_k)$ and the unknown part~$\dynlearned(\sysstate_k, \sysinput_k)$ are Lipschitz continuous with known Lipschitz constants~(see~\marginref{mrgn:lipschitz_cont})~\cite{Berkenkamp2017}.
    Given a Lyapunov function~(see~\marginref{mrgn:lyapunov}), 
    an initial safe policy~$\boldsymbol{\pi}_0$, and a GP to learn the unknown dynamics~$\dynlearned$, a control policy~$\boldsymbol{\pi}_k$ is chosen so that it maximizes a conservative estimate of the \textit{region of attraction~(ROA)}~(see~\marginref{mrgn:roa}).
The most uncertain states (based on the GP's covariance) inside the ROA, are explored. This reduces the uncertainty over time and allows the ROA to be extended. 
The practical implementation resorts to discrete states for tractability
    and retains the stability guarantees while being sub-optimal in (exploration) performance.

 \subsection{Encouraging Safety and Robustness in Reinforcement Learning}
\label{sec:methods-rl}

The approaches in this section are safety-augmented variations of the traditional MDP and RL frameworks.
In general, these methods do not assume knowledge of an \emph{a priori} nominal model $\dynprior$ and some  also learn the reward or step cost $\coststep$~\cite{Turchetta2016}, or the safety constraints~$\cstr$~\cite{dalal2018a}.
Rather than providing strict safety guarantees, these approaches encourage constraint satisfaction during and after learning, or  robustness of the learned control policy $\policy$ to uncertain dynamics (Safety Level~I, see \textbf{\autoref{fig:survey_summary}}).
In plain MDP formulations,
\emph{(i)} states and inputs (or \emph{actions}) are assumed to have known, often discrete and finite, domains but they are not further constrained while searching for an optimal policy $\policy^*$
and \emph{(ii)} only loose assumptions are made on the dynamics $\dyn$, for example, the system being Markovian~\cite{Sutton2018} (see \marginref{mrgn:markov_prop}).

\begin{marginnote}[]
\refstepcounter{margin}
\entry{M\arabic{margin}. Markov Property}{
The probability for a system to be in state $\sysstate_k$ at time $k$ only depends on the state at time $k-1$, $\sysstate_{k-1}$ and, in MDPs, $\sysinput_{k-1}$.
}
\label{mrgn:markov_prop}
\end{marginnote}

A previous taxonomy of safe RL~\cite{Garcia2015}---covering research published up until 2015---distinguished methods that either modify the exploration process to include external knowledge or modify the optimality criterion $\cost$ with a safety factor.
However, as pointed out in~\cite{drlthatmatters}, the number and breadth of publications in RL, including safe RL, has since  greatly increased.
Because the approaches recently proposed in this area are numerous and diverse,
we provide a high-level review of some of the most significant trends with a focus on their applicability to robotics. These include:
\emph{(i)}
the safe exploration of MDPs;
\emph{(ii)}
risk-aware and cautious RL adaptation;
\emph{(iii)}~RL based on constrained MDPs;
and \emph{(iv)}
robust RL.

\begin{marginnote}[]
\refstepcounter{margin}
\entry{M\arabic{margin}. Ergodicity}{
An MDP is \textit{ergodic} if, by following a given policy, any state is reachable from any other state. 
    Ergodic properties of dynamical systems are key to establish conditions for their controllability~\cite{Mezic2003}.
}
\label{mrgn:ergodicity}
\end{marginnote}
\subsubsection{Safe Exploration and Optimization}
\label{sec:methods-rl:safe-exp}

An RL algorithm's need to explore poses a challenge to its safety as it must select inputs  with unpredictable consequences in order to learn about them.

\subheading{Safe Exploration.}
\label{sec:safe-exp}
The problem of safely exploring an MDP is tackled
in~\cite{moldovan2012a} through the notion of ergodicity (see \marginref{mrgn:ergodicity}).
Policy updates that preserve the ergodicity of the MDP enable the system to return to an arbitrary state from any state.
Thus, the core idea is to restrict the space of eligible policies to those that make the MDP ergodic (with at least a given probability). Exactly solving this problem, however, is NP-hard. A simplified problem is solved in~\cite{moldovan2012a}, using a heuristic exploration algorithm~\cite{brafman2002rmax},
which leads to sub-optimal but safe exploration that only considers a subset of the ergodic policies.
In practice, \cite{moldovan2012a} is demonstrated in two simulated scenarios with discrete, finite $\set{X}$ and $\set{U}$.
Rather than designing for a recoverable exploration through ergodicity, safe exploration strategies that provide constraint satisfaction are developed in~\cite{pham2018optlayer,dalal2018a}.
A \emph{safety layer} is used to convert an optimal but potentially unsafe action $\vec{u}_{\mathrm{learn},k}$ produced by an NN policy, to the closest safe action $\vec{u}_{\mathrm{safe},k}$ with respect to some safety state constraint. Both \cite{pham2018optlayer} and \cite{dalal2018a} involve solving a constrained least squares problem, $\vec{u}_{\mathrm{safe},k} = \argmin_{\vec{u}_k} \lVert \vec{u}_k - \vec{u}_{\mathrm{learn},k} \rVert_2^2$, which is akin to the safety filter approaches further discussed in \autoref{sec:set_certification}.
Concretely, \cite{pham2018optlayer} assumes full knowledge of the (linear) constraint functions and solves $\vec{u}_{\mathrm{safe},k}$ using a differentiable Quadratic Programming (QP) solver. In contrast, \cite{dalal2018a} assumes constraints are unknown \emph{a priori} but can be evaluated. Thus, the approach learns the parameters of linear approximations to these constraints and then uses them in the solver. Notably, \cite{pham2018optlayer,dalal2018a} consider single time-step state constraints. In \autoref{sec:methods-rl:cmdps}, we will also discuss methods that deal with more general trajectory-level constraints (\autoref{eq:expected_constraints}).

\subheading{Safe Optimization.}
\label{sec:safeopt}
Several works address the problem of safely optimizing an unknown function (typically the cost function), often exploiting GP models~\cite{Kim2021}. Safety refers to sampling inputs that do not violate a given safety threshold (\autoref{eq:hard_constraints}, \autoref{eq:chance_constraints}). These approaches fall under the category of Bayesian optimization~\cite{duivenvoorden2017a} and include \emph{SafeOpt}~\cite{sui2015a}; \emph{SafeOpt-MC}~\cite{Berkenkamp2016a}, an extension towards multiple constraints; \emph{StageOpt}~\cite{sui2018a}, a the more efficient two-stage implementation; and \emph{GoSafe}~\cite{baumann2021gosafe}, to explore beyond the initial safe region.
In particular, \emph{SafeOpt} infers two subsets of the safe set from the GP model: \emph{(i)} one with candidate inputs to extend the safe set, and \emph{(ii)} one with candidate input to optimize the unknown function---from which it greedily picks the most uncertain.
The ideas pioneered by \emph{SafeOpt} were applied to MDPs in
\emph{SafeMDP}~\cite{Turchetta2016} resulting in safe exploration of MDPs with an unknown cost function $\coststep(\sysstate, \sysinput)$, which \cite{Turchetta2016} models as a GP.
    In \emph{SafeMDP}, the single-step reward represents the safety feature that should not violate a given threshold.
Another extension of~\cite{sui2015a}, \emph{SafeExpOpt-MDP}~\cite{wachi2018a}, 
treats the safety feature $c^j$ and the MDP's cost $\coststep$ as two separate, unknown functions, allowing for the constraint of the first and the optimization of the latter.
A recent survey of these techniques was provided in~\cite{Kim2021},
highlighting the distinction between safe learning in regression---i.e., minimizing the selection and evaluation of non-safe training inputs---and safe exploration in MDPs and stochastic dynamical systems like \autoref{eq:dynamics_ps}---i.e., selecting action inputs that also preserve ergodicity).

\begin{marginnote}[]
\refstepcounter{margin}
\entry{M\arabic{margin}. Conservative Q-learning (CQL)}{
    A recent offline RL approach~\cite{kumar2020cql} to mitigate the overestimation problem in value functions. Q-learning is known to overestimate action-state values because of its maximization operator, and offline RL can lead to overestimated values of the actions that are more likely under the policy used to collect the data. CQL tackles this by using a novel update rule with a simple regularizer to learn value lower bounds instead.
}
\label{mrgn:cql}
\end{marginnote}
\begin{marginnote}[]
\refstepcounter{margin}
\entry{M\arabic{margin}. Model Ensembles}{
    Collections of learned models can be used to mitigate noise or better capture the stochasticity of a system.
    \emph{Probabilistic Ensembles with Trajectory Sampling} (PETS)~\cite{chua2018a} is a model-based RL method that uses model ensembles and probabilistic networks to capture uncertainties in the system dynamics. }
\label{mrgn:pets}
\end{marginnote}
\subheading{Learning a Safety Critic.}
A safety critic is a learnable action-value function $Q^{\policy}_{\mathrm{safe}}$ that can be used to detect if a proposed action can lead to unsafe conditions.
The works~\cite{srinivasan2020learning,thananjeyan2020b,bharadhwaj2021conservative}
make use of such a critic and then resort to various fallback schemes to determine a safer alternative input.
These works differ from those in \autoref{sec:set_certification} in that the filtering criterion depends on a model-free, learned value function, which can only grant the satisfaction of Safety Level I.  

\begin{marginnote}[]
\refstepcounter{margin}
\entry{M\arabic{margin}. Conditional Value at Risk (CVaR)}{
CVaR$_{\alpha}$ is a  risk measure that is equal to the average of the worst $\alpha$-percentile of the total cost $\cost$~\cite{chow2017risk}. 
}
\label{mrgn:cvar}
\end{marginnote}

In \cite{srinivasan2020learning},  \emph{safety Q-functions for reinforcement learning} (SQRL) are used to \emph{(i)} learn a safety critic from only abstract, sparse safety labels (e.g., a binary indicator), and \emph{(ii)}~transfer knowledge of safe action inputs to new but similar tasks. 
SQRL trains $Q^{\policy}_{\mathrm{safe}}$ to predict the future probability of failure in a trajectory, and uses it to filter out unsafe actions from the policy $\policy$. 
The knowledge transfer is achieved by pre-training $Q^{\policy}_{\mathrm{safe}}$ and $\policy$ in simulations, and then fine-tuning $\policy$ on the new task (with similar dynamics $\dyn$ and safety constraints), still in simulation, while reusing $Q^{\policy}_{\mathrm{safe}}$ to discriminate unsafe inputs. 
However, the success of the final safe policy still depends on the task- and environment-specific hyperparameters~\cite{srinivasan2020learning} (which must be found \emph{via} parameter search, in simulation, prior to the actual experiment).
Building on~\cite{srinivasan2020learning}, \emph{Recovery RL}~\cite{thananjeyan2020b} additionally learns a recovery policy $\policy_{\mathrm{rec}}$ to produce fallback actions for $Q^{\policy}_{\mathrm{safe}}$ as an alternative to filtering out unsafe inputs and resorting to potentially sub-optimal ones in $\policy$. 
In~\cite{bharadhwaj2021conservative}, the authors extend Conservative Q-Learning  (CQL, see \marginref{mrgn:cql}) and propose Conservative Safety Critic (CSC). Similarly to~\cite{srinivasan2020learning}, CSC assumes sparse safety labels and uses $Q^{\policy}_{\mathrm{safe}}$ for action filtering, but it trains $Q^{\policy}_{\mathrm{safe}}$ to upper bound the probability of failure and ensures provably safe policy improvement at each iteration.

\subsubsection{Risk-Averse RL and Uncertainty-Aware RL}
\label{sec:methods-rl:risk}
Safety in  RL can also be encouraged by deriving and using risk (or uncertainty) estimates during learning.
These estimates are typically computed for the system dynamics or the overall cost function, and leveraged to produce more conservative (and safer) policies.

Risk can be defined as the probability of collision for a robot performing a navigation task~\cite{kahn2017a,lutjens2019a}.
A collision model, captured by a neural network ensemble trained with Monte Carlo dropout, predicts the probability distribution of a collision, given the current state and a sequence of future actions.
The collision-averse behavior is then achieved by incorporating the collision model in a MPC planner.

In~\cite{zhang2020cautious}, the authors build upon PETS~\cite{chua2018a} (see~\marginref{mrgn:pets}) to propose \emph{cautious adaptation for safe RL} (CARL).
CARL is composed of two training steps:
\emph{(i)} a pre-training phase that is not risk-aware, where a PETS agent is trained on multiple different system dynamics, 
and \emph{(ii)} an adaptation phase, 
where the agent is fine-tuned on the target system by taking risk-averse actions. CARL also proposed two notions of risk, one to avoid low-reward trajectories and another to avoid \emph{catastrophes} (e.g., irrecoverable states or constraint violations).
In SAVED~\cite{thananjeyan2020a}, a PETS~\cite{chua2018a} agent is used to predict the probability of a robot's collision and to evaluate a chance constraint for safe exploration.
Similarly to the methods in the last paragraph of \autoref{sec:methods-rl:safe-exp}, SAVED also learns a value function from sparse costs, which it uses as a terminal cost estimate. 

To learn risk-averse policies using only offline data, the approach in \cite{urpi2021riskaverse} optimizes for a risk measure of the cost such as Conditional Value-at-Risk (CVaR) (see \marginref{mrgn:cvar}). 
Instead of using model ensembles as in \cite{kahn2017a,lutjens2019a,zhang2020cautious}, the work in \cite{urpi2021riskaverse} uses distributional RL (see \marginref{mrgn:distr_rl}) to explicitly model the distribution of the total cost of the task (control of a simulated 1D car), and offline learning to improve scalability.

\subsubsection{Constrained MDPs and RL}
\label{sec:methods-rl:cmdps}

The CMDP framework is frequently used in safe RL, as it introduces constraints that can be used to express arbitrary safety notions
in the form of \autoref{eq:expected_constraints}.
RL for CMDPs, however, faces two important challenges (see~\autoref{subsec:reinforcement_learning_background}): \textit{(i)} how to incorporate and enforce constraints in the RL algorithm; and \textit{(ii)} how to efficiently solve the constrained RL problem---especially when using deep learning models, which are the \emph{de facto} standard in non-safe RL. 
Works in this section that address these questions include \textit{(i)}~Lagrangian methods for RL, \textit{(ii)}~generalized Lyapunov functions for constraints, and \textit{(iii)}~{Backward Value Functions}.
Nonetheless, much of the work in this section remains confined to rather naive simulated tasks, motivating further investigation into the applicability of constrained RL in real-world control. 

\begin{marginnote}[]
\refstepcounter{margin}
\entry{M\arabic{margin}. Distributional RL}{
Distributional RL~\cite{bellemare2017} focuses on the modeling and learning of the \emph{distribution} of the cost $\cost$,
    rather than its expected value. The greater descriptive power can enable richer predictions and risk-aware behaviours. 
}
\label{mrgn:distr_rl}
\end{marginnote}

\begin{marginnote}[]
\refstepcounter{margin}
\entry{M\arabic{margin}. Lagrangian Methods}{
    To solve a constrained optimization problem $\min_x f(x)\ \text{s.t.}\ g(x) \le 0$, \textit{Lagrangian methods} define a Lagrangian function $\mathcal{L} (x, \lambda) = f(x) + \lambda g(x)$ on primal variable $x$, dual variable $\lambda$, and solve the equivalent unconstrained problem $\max_{\lambda \ge 0} \min_x \mathcal{L} (x, \lambda)$. In practice, numerical approaches, such as gradient methods, are often used to perform the primal-dual updates~\cite{chow2017risk}. 
}
\label{mrgn:lagrangian}
\end{marginnote}

\subheading{Lagrangian Methods in RL Optimization.} In \cite{chow2017risk,liang2018accelerated}, the CMDP optimization problem (see~\autoref{subsec:reinforcement_learning_background}) is first transformed into an 
unconstrained optimization one (see \marginref{mrgn:lagrangian}) over the primal variable $\policy$ and dual variable $\lambda$ using the Lagrangian $\mathcal{L} (\policy, \lambda)$, and RL is used as a subroutine in the primal-dual updates for \autoref{eq:lagrange_cmdp_update}:
\begin{subequations}
    \label{eq:lagrange_cmdp}
    \begin{gather}
    \mathcal{L} (\policy, \lambda) = J^{\policy} + \sum_{j} \lambda_j \left( J^{\policy}_{c^j} - d^j \right) \\ 
(\policy^*, \lambda^*) = \arg \max_{\lambda \ge 0} \min_{\policy} \mathcal{L} (\policy, \lambda)  \label{eq:lagrange_cmdp_update} \quad \text{s.t.}\quad \text{\autoref{eq:dynamics_ps}, \ref{eq:policy_ps}, \ref{eq:initial_state}}\,.
    \end{gather}
\end{subequations}
In particular, \cite{chow2017risk} defines a constraint on the CVaR of cost $J^{\policy}$, and uses policy gradient or actor-critic methods to update the policy in \autoref{eq:lagrange_cmdp_update}.
In subsequent work~\cite{liang2018accelerated}, the authors improve upon \cite{chow2017risk} by incorporating off-policy updates of the dual variable (with the on-policy primal-dual updates). This is empirically shown to achieve better sample efficiency and faster convergence. 
In \cite{achiam2017constrained}, the authors extend a standard trust-region (see~\marginref{mrgn:divergence}) RL algorithm~\cite{schulman2015a} 
to CMDPs using a novel bound that relates the expected cost of two policies to their state-averaged divergence (see~\marginref{mrgn:divergence}).
The key idea is performing primal-dual updates with surrogates/approximations of the cost $J^{\policy}$ and constraint cost $J^{\policy}_{c^j}$ derived from the bound.
The benefits are two-fold: \textit{(i)} the surrogates can be estimated with only state-action data, bypassing the challenge of trajectory evaluation from off-policy data; and \textit{(ii)} the updates guarantee monotonic policy improvement and near constraint satisfaction at each iteration (Safety Level I). However, unlike \cite{chow2017risk, liang2018accelerated}, each update involves solving the dual variables from scratch, which can be computationally expensive.

\subheading{A Lyapunov Approach to Safe RL.} 
Lyapunov functions (see~\marginref{mrgn:lyapunov}) are used extensively in control to analyze system stability and they are a powerful tool to translate a system's global properties into local conditions. 
In \cite{chow2018lyapunov,chow2019lyapunov}, Lyapunov functions are used to transform the trajectory-level constraints $\cost^{\policy}_{c_j}$ in \autoref{eq:expected_constraints}
into step-wise, state-based constraints. This allows a more efficient computation of $\cost^{\policy}_{c_j}$ and mitigates the cost of off-policy evaluation. 
These approaches, however, require the system to start from a baseline policy $\policy_0$ that already satisfies the constraints.
In \cite{chow2018lyapunov}, four different algorithms are proposed to
solve CMDPs by combining traditional RL methods and the Lyapunov constraints, but they are only applicable to discrete input spaces (with continuous state spaces).  
In subsequent work~\cite{chow2019lyapunov}, the authors of~\cite{chow2018lyapunov} extend the approach to continuous input spaces and standard policy gradient methods,
addressing its computational tractability.

\subheading{Learning Backward Value Functions.}
The work in~\cite{satija2020constrained} proposes Backward Value Functions (BVF)
as a way to overcome the excessive computational cost of the approaches in the previous paragraphs~\cite{achiam2017constrained,chow2019lyapunov}.
Similarly to a (forward) value function $V^{\policy}$ that estimates the total future cost from each state, a BVF ${V}^{b, \policy}$ estimates the accumulated cost so far (up to the current state). We can decompose a trajectory-level constraint at any time step $k$ as sum of
$V^{\policy}_{c^j}(\sysstate_k)$ and ${V}^{b,\policy}_{c^j}(\sysstate_k)$
for the constraint cost $J^{\policy}_{c^j}$. 
This decomposition also alleviates the problem of off-policy evaluation as these value functions can be learned concurrently and efficiently via Temporal Difference (TD) methods (see \marginref{mrgn:td_learning}). 
In practice,  $V^{\policy}_{c^j}$, ${V}^{b,\policy}_{c^j}$, $V^{\policy}$ are jointly learned \cite{satija2020constrained} and used for policy improvement at each time step, allowing to implement Safety Level I.
The approach is intended for discrete action spaces but can be adapted to continuous ones as in~\cite{dalal2018a}. 

\begin{marginnote}[]
\refstepcounter{margin}
\entry{M\arabic{margin}. Divergence}{
    \textit{Divergence} quantifies the similarity between probability distributions. It may not satisfy the symmetry and the triangle inequality of a distance metric. In RL, commonly used ones include Kullback-Leibler divergence (KL), total variation distance (TV), and the Wasserstein distance, which is also a valid distance metric.
    Divergence measures can be used to constrain policy updates in trust-region RL algorithms~\cite{schulman2015a}.
}
\label{mrgn:divergence}
\end{marginnote}

\subsubsection{Robust MDPs and RL}
\label{sec:methods-rl:rrl}

Works in this section aim to implement robustness in RL, specifically, learning policies that can operate under disturbances and generalize across similar tasks or robotic systems.
This is typically done by framing the learning problem as a robust MDP (\autoref{eq:minimax_rl}).
In~\cite{morimoto2005}, Robust RL (RRL) implements an Actor-Disturber-Critic architecture. The authors observes that the learned policy and value function coincide with those derived analytically from $\Hfunc_{\infty}$ control theory for linear systems (see \autoref{subsec:control_background}).
However, more recent robust RL literature often abstains from assumptions on dynamics or disturbances, and applies model-free RL~\cite{turchetta2020robustmodelfreerl} to seek empirically robust performance at the expense of theoretical guarantees. 
In the following paragraphs, we introduce two lines of work: \textit{(i)} robust adversarial RL, which explicitly models the min-max problem in~\autoref{eq:minimax_rl} in a game theoretic fashion; and \textit{(ii)}~domain randomization, which approximates the same problem in~\autoref{eq:minimax_rl} by learning on a set of randomized perturbed dynamics.

\begin{marginnote}[]
\refstepcounter{margin}
\entry{M\arabic{margin}. Temporal Difference Learning}{
    \textit{Temporal Difference (TD) learning} \cite{Sutton2018} is an important class of model-free RL methods that updates the value function by bootstrapping from the current estimate using the Bellman equation to formulate sampling-based, iterative updates.
    For example, given a control step data tuple $(\sysstate_k, l_k, \sysstate_{k+1})$ we can perform a learning step as $V^{\policy}(\sysstate_k) \leftarrow V^{\policy}(\sysstate_k) + \alpha (l_k + \gamma V^{\policy}(\sysstate_{k+1}) - V^{\policy}(\sysstate_k))$ where $\alpha, \gamma$ are the step size and discount factor.
}
\label{mrgn:td_learning}
\end{marginnote}

\subheading{Robustness Through Adversarial Training.}
Combining RL with adversarial learning~\cite{goodfellow2014gan} results in robust adversarial RL (RARL)~\cite{pinto2017robust},
where the robust optimization problem (\autoref{eq:minimax_rl}) is set up as a two-player, discounted zero-sum Markov game in which an agent (protagonist) learns policy $\policy$ to control the system and another agent (adversary) learns a separate policy to destabilize the system.
The two agents learn in an alternated fashion (each is updated while fixing the other), attempting to progressively improve both the robustness of the protagonist's policy and the strength of its adversary.
The work of~\cite{pan2019a} extends~\cite{pinto2017robust} with risk-aware agents (see \autoref{sec:methods-rl:risk}), with the protagonist being risk-averse and the adversary being risk-seeking.
This method learns an ensemble of Deep Q-Networks (DQN)~\cite{mnih2015a} and defines the risk of an action based on the variance of its value predictions.
In another extension of~\cite{pinto2017robust}, a population of adversaries (rather than a single one) is trained~\cite{vinitsky2020robust},
leading to the resulting protagonist being less exploitable by new adversaries.
Finally, the work in \cite{everett2020certified}  
proposes certified lower bounds for the value predictions from a DQN~\cite{mnih2015a}, given bounded observation perturbations.
The action selection is based on these value lower bounds, assuming adversarial perturbation.

\subheading{Robustness Through Domain Randomization.}
Domain randomization methods attempt to learn policies that empirically generalize to a wider range of tasks or robot systems.
Instead of considering worst-case disturbances or scenarios, learning happens on systems with randomly perturbed parameters (e.g., inertial properties, friction coefficients).
These parameters often have pre-specified ranges, effectively inducing a robust set that \textit{(i)} approximates the uncertainty set $\set{D}$ in~\autoref{eq:minimax_rl}, and \textit{(ii)} allows the system to reuse any standard RL algorithm. 
As an example, in~\cite{sadeghi2017card2rl}, a quadrotor learns vision-based flight  purely in simulation from scenes with randomized properties. Using this model-free policy in the real world resulted in improved 
collision avoidance performance. Instead of learning a policy directly, the work in~\cite{loquercio2019a} uses  learned visual predictions with a MPC controller to enable efficient and scalable real-world performance.
Besides the uniform randomization in~\cite{sadeghi2017card2rl, loquercio2019a}, adaptive randomization strategies such as
Bayesian search~\cite{rajeswaran2017epopt} are also a promising direction. In \cite{mehta20a}, a discriminator is adversarially trained and used to guide the randomization process that generates systems that are less explored or exploited by the current policy.

\subsection{Certifying Learning-Based Control Under Dynamics Uncertainty}
\label{sec:certification}
In this section, we review methods providing certification to learning-based control approaches that do not inherently account for  safety constraints, see \textbf{\autoref{fig:block_diagram}}. We divide the discussion into two parts: \textit{(i)} stability certification, and \textit{(ii)} constraint set certification. The works in this section leverage an \textit{a priori} dynamics model of the system and provide hard or probabilistic safety guarantees~(Safety Level II or III) under dynamics uncertainties~(see \textbf{\autoref{fig:survey_summary}}).
\subsubsection{Stability Certification} \label{sec:stability_certification}
This section introduces certification approaches that guarantee closed-loop stability under a learning-based control policy. 

\subheading{Lipschitz-Based Safety Certification for DNN-Based Learning Controllers.}
These approaches exploit the expressive power of DNNs  for policy parametrization and guarantee closed-loop stability through a Lipschitz constraint on the DNN. Let $\rho$ be a Lipschitz constant of a DNN policy (see~\marginref{mrgn:lipschitz_cont}),
an upper bound on the Lipschitz constant~$\rho$ that guarantees closed-loop stability (Safety Level~III) can be established 
by using a small-gain stability analysis~\cite{zhou2020b}, solving a semi-definite program (SDP)~\citep{jin2020stability}, or applying a sliding mode control framework~\citep{shi2019a}. This bound on~$\rho$ can be used in either \textit{(i)} a passive, iterative enforcement approach where the Lipschitz constant~$\rho$ is first estimated (e.g., via an SDP-based estimation~\citep{fazlyab2019efficient}) and then used to guide re-training until the Lipschitz constraint is satisfied, or \textit{(ii)} an active enforcement approach where the Lipschitz constraint is directly enforced by the training algorithm of the DNN (e.g., via spectral normalization~\citep{shi2019a}). While guaranteeing stability, the Lipschitz-based certification approaches often rely on the particular structure of the system dynamics (e.g., a control-affine structure or linear structure with additive nonlinear uncertainty) to find the certifying Lipschitz constant. It remains to be explored how this idea can be extended to more generic robot systems.

\subheading{Learning Regions of Attraction for Safety Certification.}
The region of attraction (ROA) of a closed-loop system (see~\marginref{mrgn:roa}) is used in the learning-based control literature as a means to guarantee safety. For a nonlinear system with a given state-feedback controller, the ROA is the set of states  that is guaranteed to converge to the equilibrium, which is treated as a safe state.
This notion of safety provides a mean to certify a learning-based controller. It guarantees that there is a region in state space from which the controller
can drive the system back to the safe state (Safety Level~III)~\citep{Richards2018a}. ROAs can, for example, be used to guide data acquisition for model or controller learning~\citep{Berkenkamp2017,zhou2020general}. 

We consider deterministic closed-loop systems $\sysstate_{k+1} = \dyn_{\policy}(\sysstate_k) = \dyn(\sysstate_k,\policy(\sysstate_k))$ with $\dyn_{\policy}(\sysstate_k)$ being Lipschitz continuous. A Lyapunov neural network (LNN) can be used  to iteratively learn the ROA of a controlled nonlinear system from the system's input-output data~\citep{Richards2018a}. As compared to the typical Lyapunov functions in control (e.g., quadratic Lyapunov functions), the proposed method uses the LNN as a more flexible Lyapunov function representation to provide a less conservative estimate of the system's ROA. The necessary properties of a Lyapunov function are preserved via the network's architectural design. In~\citep{zhou2020general},  an ROA estimation approach for high-dimensional systems is presented, combining a sum of squares (SOS) programming method for the ROA computation~\citep{jarvis-wloszek2003sos} and a dynamics model order reduction technique to curtail computational complexity~\citep{schilders2008model}.

\subsubsection{Constraint Set Certification}
\label{sec:set_certification}
This section summarizes approaches that provide constraint set certification to a learning-based controller based on the notion of  \textit{robust positive control invariant} (RPCI) safe sets~$\Omega_{\mathrm{safe}}\subseteq \stateconstraints$~(see~\marginref{mrgn:r_pos_ctrl_inv_set}).\begin{marginnote}[]
\refstepcounter{margin}
\entry{M\arabic{margin}. Robust Positive Control Invariant Safe Set}{
A \textit{robust positive control invariant} (RPCI) safe set~$\Omega_\mathrm{safe}$ is a set contained in $\stateconstraints$ such that, if we start within the safe set, there exists a feedback policy $\pi(\vec{x})$ to keep the system in the safe set despite all possible model errors and process noise, captured by $\set{D}$.
}
\label{mrgn:r_pos_ctrl_inv_set}
\end{marginnote}Certified learning, which can be achieved through a safety filter~\cite{Hewing2020} or shielding~\cite{Alshiekh2018}, finds the minimal modification of a learning-based control input~$\sysinput_{\mathrm{learn}}$ (see \textbf{\autoref{fig:block_diagram}}) such that the system’s state stays inside the set~$\Omega_{\mathrm{safe}}$:
\begin{subequations}
\label{eq:safety_certification}
\begin{align}
        \vec{u}_{\mathrm{safe},k} = \argmin_{\sysinput_k\in\inputconstraints} &\quad \lVert \sysinput_k - \sysinput_{\mathrm{learn},k} \rVert_2^2 \label{subeq:safety_filter_objective}\\
        \mathrm{s.t.} &\quad \vec{x}_{k+1} = \dynprior_k(\sysstate_{k}, \sysinput_{k}) + \dynlearned_k(\sysstate_{k}, \sysinput_{k}, \vec{w}_{k}) \in \Omega_{\mathrm{safe}},\label{subeq:safety_filter_constraint} \\ &\quad\forall\ \dynlearned_k(\sysstate_{k}, \sysinput_{k}, \vec{w}_{k})\in\set{D}(\sysstate_k, \sysinput_k)\nonumber  \,, 
\end{align}
\end{subequations}
where the range of possible disturbances  $\set{D}(\sysstate_k, \sysinput_k)$ is given.
Since the safety filter and controller are usually decoupled, suboptimal behavior can emerge as the learning-based controller may try to violate the constraints~\cite{Koller2019b}.

\subheading{Control Barrier Functions (CBFs).}
CBFs are used to define safe sets. More specifically, the safe set $\Omega_{\mathrm{safe}}$ is defined as the superlevel set of a continuously differentiable CBF~$B_{\mathrm{c}}$, $B_{\mathrm{c}} : \R^{n_x} \to \R$, as 
$\Omega_{\mathrm{safe}} = \{ \sysstate \in \R^{n_x} : B_{\mathrm{c}}(\sysstate) \geq 0 \}$. The function $B_{\mathrm{c}}$ is generally considered for continuous-time, control-affine systems~(see also~\marginref{mrgn:cont_time}) of the form:
\begin{equation}
\label{eq:control_affine}
 \dot{\sysstate} = \dyn_{\sysstate}(\sysstate) + \dyn_{\sysinput}(\sysstate)\sysinput\,,   
\end{equation}
where $\dyn_{\sysstate}$ and $\dyn_{\sysinput}$ are locally Lipschitz and $\sysstate, \sysinput$ are functions of time~\cite{ames2019a}.
In the simplest form, the function $B_{\mathrm{c}}$ is a CBF if 
there exists
a state-feedback control input $\sysinput$, such that the time derivative $\dot{B}_{\mathrm{c}}(\sysstate) = \frac{\partial B_{\mathrm{c}}}{\partial \sysstate} \dot{\sysstate}$ 
satisfies~\cite{ames2019a}: 
\begin{equation}
\label{eq:cbf_condition}
\sup_{\sysinput \in \set{U}} \frac{\partial B_{\mathrm{c}}}{\partial \sysstate} \left( \dyn_{\sysstate}(\sysstate) + \dyn_{\sysinput}(\sysstate)\sysinput \right) \geq - B_{\mathrm{c}}(\sysstate) \,.
\end{equation}
The CBF condition in \autoref{eq:cbf_condition} is a continuous-time version of the robust positive control invariance constraint in \autoref{subeq:safety_filter_constraint}.\begin{marginnote}[]
\refstepcounter{margin}
\entry{M\arabic{margin}. Control Lyapunov Functions}{
\textit{Control Lyapunov Functions} (CLFs) extend the notion of Lyapunov stability to continuous-time, control-affine systems (\autoref{eq:control_affine}). A positive definite function $\lyapunov_{\mathrm{c}}$ is a CLF, if there exists a state-feedback control input $\sysinput$, such that the time derivative $\dot{\lyapunov}_{\mathrm{c}}$ satisfies: $\dot{\lyapunov}_{\mathrm{c}}(\sysstate) \leq - \lyapunov_{\mathrm{c}}(\sysstate)$. 
The existence of a CLF guarantees that there also exists a state-feedback controller~$\policy(\vec{x})$ that asymptotically stabilizes the system. 
}
\label{mrgn:clf}
\end{marginnote}In addition to the RPCI constraint, a similar constraint for asymptotic stability can be added to \autoref{eq:safety_certification} in the form of a constraint on the time derivative of a control Lyapunov function (CLF, see~\marginref{mrgn:clf}) $\lyapunov_{\mathrm{c}}$.
However, uncertain dynamics also yield uncertain time derivatives of~$B_{\mathrm{c}}$ and~$\lyapunov_{\mathrm{c}}$.

Learning-based approaches extend CBF and CLF analyses to control-affine systems in \autoref{eq:control_affine} with known nominal $\dynprior_{\sysstate}$, $\dynprior_{\sysinput}$ and unknown $\dynlearned_{\sysstate}$, $\dynlearned_{\sysinput}$.  
The time derivative of the unknown dynamics 
for CLFs and/or CBFs (e.g., $\frac{\partial B_{\mathrm{c}}}{\partial \sysstate} (\dynlearned_{\sysstate}(\sysstate) + \dynlearned_{\sysinput}(\sysstate)\sysinput)$)  can be learned from iterative trials~\cite{taylor2019a, taylor2020a} or data collected by an RL agent~\cite{Ohnishi2019, Choi2020}.
Given a CBF and/or CLF for the true dynamics $\dyn$, improving the estimate of the CBF's or CLF's time derivative for the unknown dynamics $\dynlearned$ using data, collected either offline or online, yields a more precise estimate of the constraint in \autoref{eq:cbf_condition}. However, any learning error in the CBF's or CLF's time derivative of the unknown dynamics can still lead to applying falsely certified control inputs. 
To this end, ideas from robust control can be used to guarantee set invariance~(Safety Level III) during the learning process by mapping a bounded uncertainty in the dynamics to a bounded uncertainty in the time derivatives of the CBF or CLF~\cite{taylor2019b, taylor2020b}, or by accounting for all model errors consistent with the collected data~\cite{taylor2020d}. 
In addition, adaptive control approaches have also been proposed to allow safe adaptation of parametric uncertainties in the time derivatives of the CBF or CLF~\cite{taylor2020c,Lopez2020}. 
Probabilistic learning techniques for CBFs and CLFs have been used to achieve set invariance probabilistically~(Safety Level II) with varying assumptions on the system dynamics: the function~$\dyn_{\sysinput}(\sysstate)$ is fully known~\cite{cheng2019a,Fan2019}, a nominal model is known~\cite{Wang2018a}, and no nominal model is available~\cite{khojasteh2020a}.

\subheading{Hamilton-Jacobi Reachability Analysis.} 
Another approach for state constraint set certification of a learning-based controller is via the Hamilton-Jacobi (HJ) reachability analysis. The HJ reachability analysis provides a means to estimate a robust positive control invariant safe set~$\Omega_\text{safe}$ under dynamics uncertainties (see~\marginref{mrgn:r_pos_ctrl_inv_set} and \autoref{eq:safety_certification}). 
Consider a nonlinear system subject to unknown but bounded disturbances $\dynlearned(\vec{x})\in \set{D}(\vec{x})$, where $\set{D}(\vec{x})$ is assumed to be known but possibly conservative.
To compute $\Omega_{\mathrm{safe}}$, a two-player, zero-sum differential game is formulated:
\begin{equation}
    \label{eq:zero_sum_game}
    V(\sysstate) = \max_{\sysinput_\text{sig}\in \set{U}_\text{sig}} \min_{\dynlearned_\text{sig}\in\set{D}_\text{sig}} \left(\inf_{k\ge 0} l_\text{c}\left(\phi(\sysstate, k; \sysinput_\text{sig}, \dynlearned_\text{sig})\right)\right),
\end{equation}
where $V$ is the value function associated with a point $\vec{x}\in\set{X}$, $l_\text{c}:\set{X} \mapsto \set{R}$ is a cost function that is non-negative for $\vec{x}\in\stateconstraints$ and negative otherwise, 
$\phi(\sysstate, k; \sysinput_\text{sig}, \dynlearned_\text{sig})$ denotes the state at $k$ along a trajectory initialized at $\vec{x}$ following input signal~$\sysinput_\text{sig}$ and disturbance signal~$\dynlearned_\text{sig}$, and $\set{U}_\text{sig}$ and $\set{D}_\text{sig}$ are collections of input and disturbance signals such that each time instance is in $\set{U}$ and $\set{D}$, respectively.
The value function $V$ can be found as the unique viscosity solution of the Hamilton-Jacobi Isaacs (HJI) variational inequality~\citep{mitchell2005time}. The safe set is then  $\Omega_{\mathrm{safe}} =\{\vec{x}\in\set{X} \:|\: V(\vec{x}) \ge 0\}$. 
Based on this formulation, we can also obtain an optimally safe policy~$\policy^*_\text{safe}$ 
that maximally steers the system towards the safe set~$\Omega_\text{safe}$ (i.e., in the greatest ascent direction of~$V$). The HJ reachability analysis allows us to define a safety filter for learning-based control approaches to guarantee constraint set satisfaction (Safety Level~II or III). In particular, given $\Omega_\text{safe}$ and $\policy^*_\text{safe}$, one can safely learn in the interior of $\Omega_\text{safe}$ and apply the optimally safe policy~$\policy^*_\text{safe}$ if the system reaches the boundary of~$\Omega_\text{safe}$. To reduce the conservativeness of the approach, a GP-based learning scheme is proposed in~\citep{Fisac2019} to adapt (and shrink) the unknown dynamics set $\set{D}(\vec{x})$ based on observed data.

The general HJ reachability analysis framework~\citep{mitchell2005time} has also been combined with online dynamics model learning for a target tracking task~\citep{Gillula2012}, with online planning for safe exploration~\citep{bajcsy19efficient}, and with temporal difference algorithms for safe RL~\citep{Fisac2019a}. In ~\citep{choi2021robust}, HJ reachability analysis and CBFs have been integrated to compute smoother control policies while circumventing the need to hand-design appropriate CBFs. Another recent extension~\citep{herbert2021scalable} proposed modifications that improve the scalability of the HJ safety analysis approach for higher-dimensional systems and was demonstrated on a ten-dimensional quadrotor tracking problem.

\subheading{Predictive Safety Filters.} 
Predictive safety filters can augment any learning-based controller to enforce state constraints, $\vec{x} \in \mathbb{X}_c$, and input constraints, $\vec{u} \in \mathbb{U}_c$. They do this by defining the safe invariant set $\Omega_{\mathrm{safe}}$ in \autoref{subeq:safety_filter_constraint} as the set of states (at the next time step) where a sequence of safe control inputs (e.g., from a backup controller) exists that allows the return to \textit{(i)} a terminal safe set $\set{X}_{\mathrm{term}}$ (see \marginref{mrgn:terminal_safe_set}), or \textit{(ii)} to previously visited safe states. 

\textbf{Model Predictive Safety Certification (MPSC)} uses the theory of robust MPC in \autoref{subsec:control_background} and learning-based robust MPC in \autoref{sec:learning_mpc} to filter the output of any learning-based controller, such as of an RL method, to ensure robust constraint satisfaction. The simplest implementation of MPSC  \cite{Wabersich2018a} uses tube-based MPC and considers the constraints in Equations~\ref{eq:dynamics_rmpc}--\ref{eq:initial_state_rmpc} but replaces the cost in \autoref{eq:objective_rmpc} with the cost in \autoref{subeq:safety_filter_objective} to find the closest input $\sysinput_k$ to the learned input $\sysinput_{\mathrm{learn},k}$ at the current time step that guarantees that we will continue to satisfy state and input constraints in the future. The main difference between MPSC and learning-based robust MPC described in \autoref{sec:learning_mpc} is that the terminal safe set $\set{X}_{\mathrm{term}}$ in \autoref{eq:initial_state_rmpc} is not coupled with the selection of the cost function in \autoref{eq:objective_rmpc}. Instead, the terminal safe set is conservatively initialized with $\set{X}_{\mathrm{term}} = \Omega_{\mathrm{tube}}$ and can grow to include state trajectories from previous iterations. This approach has been extended to probabilistic constraints by considering a probabilistic tube $\Omega_{\mathrm{tube}}$ in \cite{Wabersich2019}, and to nonlinear nominal models in \cite{Wabersich2021} (Safety Level II).

\begin{marginnote}[]
\refstepcounter{margin}
\entry{M\arabic{margin}. Terminal Safe Set}{
A \textit{terminal safe set}, denoted by~$\set{X}_{\mathrm{term}}$, is a subset of the safe state constraint space wherein a known auxiliary controller is guaranteed to preserve the system's state. Entering the terminal safe set at some fixed horizon $H$ is enforced as a constraint.
}
\label{mrgn:terminal_safe_set}
\end{marginnote}

\textbf{Backup control for safe exploration} ensures hard state constraint satisfaction~(Safety Level III) by finding a safe backup controller for any given RL policy ~$\policy$~\cite{mannucci2018a}. Under the assumptions of a known bound~$\set{D}(\sysstate, \sysinput)$ on the dynamics~$\dyn$ and a distance measure to the state constraints~$\set{X}_c$, the backup controller is used to obtain a future state in the neighborhood of a previously visited safe state in some prediction horizon. Before a control input $\sysinput_k$ from $\policy$ is applied to the system, all possible predicted states $\sysstate_{k + 1}$ must satisfy $\emph{(i)}$ $\sysstate_{k + 1} \in \set{X}_c$ and $\emph{(ii)}$ the existence of a safe backup action $\sysinput_{\mathrm{certified}}(\sysstate_{k + 1})$. Otherwise, the previous backup control input $\sysinput_{\mathrm{certified}}(\sysstate_{k})$ is applied. This procedure guarantees that the system state stays inside a robust positive control invariant set~$\Omega_{\mathrm{safe}} \subseteq \stateconstraints$.

\pgfdeclarelayer{background}
\pgfdeclarelayer{foreground}
\pgfsetlayers{background,main,foreground}

\pgfdeclarepattern{
  name=hatch,
  parameters={\hatchsize,\hatchangle,\hatchlinewidth},
  bottom left={\pgfpoint{-.1pt}{-.1pt}},
  top right={\pgfpoint{\hatchsize+.1pt}{\hatchsize+.1pt}},
  tile size={\pgfpoint{\hatchsize}{\hatchsize}},
  tile transformation={\pgftransformrotate{\hatchangle}},
  code={
    \pgfsetlinewidth{\hatchlinewidth}
    \pgfpathmoveto{\pgfpoint{-.1pt}{-.1pt}}
    \pgfpathlineto{\pgfpoint{\hatchsize+.1pt}{\hatchsize+.1pt}}
    \pgfpathmoveto{\pgfpoint{-.1pt}{\hatchsize+.1pt}}
    \pgfpathlineto{\pgfpoint{\hatchsize+.1pt}{-.1pt}}
    \pgfusepath{stroke}
    }
}

\tikzset{
hatch size/.store in=\hatchsize,
hatch angle/.store in=\hatchangle,
hatch line width/.store in=\hatchlinewidth, hatch size=5pt,
hatch angle=0pt,
hatch line width=.5pt,
}

\tikzstyle{chart}=[
    legend label/.style={font={\scriptsize},anchor=west,align=left},
    legend box/.style={rectangle, draw, minimum size=5pt},
    axis/.style={black,semithick,->},
    axis label/.style={anchor=east,font={\tiny}},
]

\tikzstyle{pie chart}=[
    chart,
    slice/.style={line cap=round, line join=round, very thick,draw=white},
    pie title/.style={font={\bf \scriptsize}},
    slice type/.style 2 args={
        ##1/.style={fill=##2},
        values of ##1/.style={}
    }
]

\newcommand{\pie}[3][]{
    \begin{scope}[#1]
    \pgfmathsetmacro{\curA}{90}
    \pgfmathsetmacro{\r}{1}
    \def\c{(0,0)}
    \node[pie title, text width=3.5cm] at (180:3.25) {#2};
    \foreach \v/\s in{#3}{
        \pgfmathsetmacro{\deltaA}{\v/100*360}
        \pgfmathsetmacro{\nextA}{\curA + \deltaA}
        \pgfmathsetmacro{\midA}{(\curA+\nextA)/2}
        \path[slice,\s] \c
            -- +(\curA:\r)
            arc (\curA:\nextA:\r)
            -- cycle;
        \pgfmathsetmacro{\d}{max((\deltaA * -(.5/50) + 1) , .5)}
        \begin{pgfonlayer}{foreground}
\end{pgfonlayer}
        \global\let\curA\nextA
    }
    \end{scope}
}

\newcommand{\legend}[2][]{
    \begin{scope}[#1]
    \path
        \foreach \n/\s in {#2}
            {
++(-105pt,0) node[\s,legend box] {} +(5pt,0) node[legend label,text width=2.25cm] {\tiny \n}
            }
    ;
    \end{scope}
}

\begin{figure}
\centering
\includegraphics[]{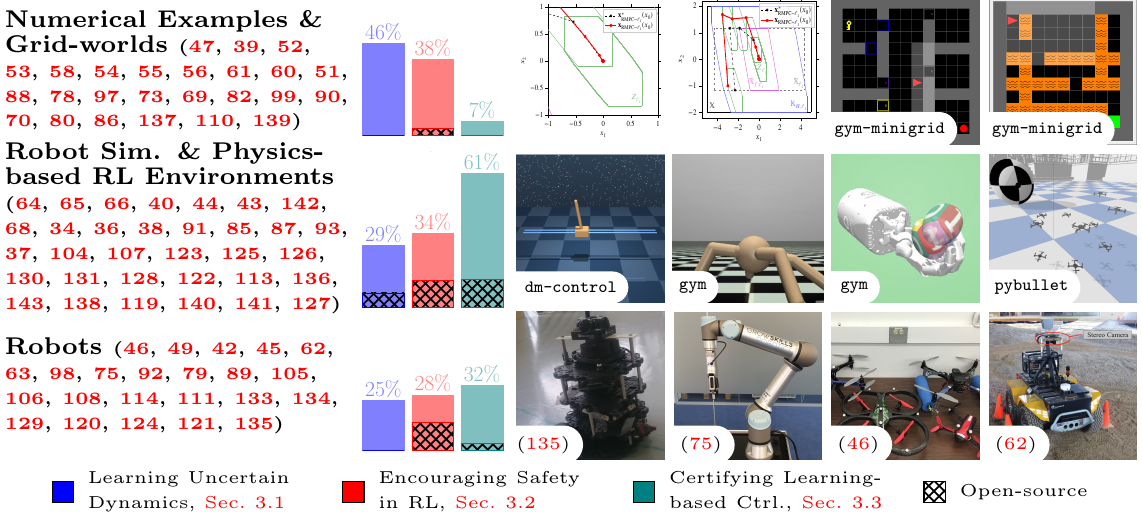}
\caption{
Summary of the environments used for evaluation.
With increasing complexity, they can be classified as: abstract numerical examples, robot simulations, and real-world robot experiments. The histograms show the prevalence of each in 
\autoref{sec:uncertain-dyn}, \autoref{sec:methods-rl}, and \autoref{sec:certification}, as well as the fraction of those whose code is open-sourced.
}
\label{fig:envs}
\end{figure}

\section{BENCHMARKS} \label{sec:benchmarks}

The approaches presented in \autoref{sec:methods} have been evaluated in vastly different ways, see  \textbf{\autoref{fig:envs}}.
The trends we observe are: The works learning uncertain dynamics (\autoref{sec:uncertain-dyn}) include a preponderance of abstract numerical examples; encouraging safety in RL (\autoref{sec:methods-rl}) mostly leverages simulations---often based on physics engines~\cite{liu2021a} like MuJoCo---but grid worlds are also common; and works certifying learning-based control (\autoref{sec:certification}) are still mostly simulated but also has the largest fraction of real-world experiments.
While numerical examples make it difficult to gauge the practical applicability of a method, we also note that even many RL environments, including physics-based ones, are not representative of existing robotic platforms--for example, they often are deterministic and do not account for variations in the environment.

In an ideal world, all research would be demonstrated in simulations that closely resemble the target system---and brought to (ideally different) real robots, whenever possible.
Furthermore, only a minority of the published research open-sourced their software implementations.
Even in RL, where sharing code is more common (see, e.g., red bars in  \textbf{\autoref{fig:envs}})---and standard environments and interfaces such as \texttt{gym}~\cite{brockman2016a} have been proposed---the reproducibility of results (that often rely on careful hyper-parameter tuning) remains challenging~\cite{drlthatmatters}.
With regard to safety, simple RL environments augmented with constraint evaluation~\cite{ray2019a} and disturbances~\cite{dulacarnold2020a} have been proposed, but lack a unified simulation  interface for both safe RL and learning-based control approaches---notably, one that also exposes the available \emph{a priori} knowledge of a dynamical system.

We believe that a necessary stepping stone for the advancement of safe learning control is to create physics-based environments that are \emph{(i)} simple enough to promote adoption, \emph{(ii)} realistic enough to represent meaningful robotic platforms, \emph{(iii)} equipped with intuitive interfaces for both control and RL researchers, and \emph{(iv)} provide useful comparison metrics (e.g., the amount of data required by different approaches). 

\subheading{Cart-Pole and Quadrotor Benchmark Environments.}
For this reason, we have created an open-source benchmark suite\footnote{Safe control benchmark suite on GitHub: \url{https://github.com/utiasDSL/safe-control-gym}}
simulating two popular platforms for the evaluation of control and RL research:
\emph{(i)} the cart-pole~\cite{kamthe2018a,everett2020certified,Fisac2019a} and \emph{(ii)} a quadrotor~\cite{mannucci2018a,Fan2020,joshi2019deep,Wang2018a,panerati2021learning}.
Our simulation environments are based on the open-source Bullet physics engine~\cite{coumans2021} and we adopt OpenAI's \texttt{gym}  interface (API)~\cite{brockman2016a} for seamless integration with many of the current RL libraries.
What sets apart our implementation from previous attempts to create safety-aware RL environments~\cite{ray2019a,dulacarnold2020a} is the extension of the traditional API~\cite{brockman2016a} with features to facilitate: \emph{(i)} the evaluation of safe learning control approaches (such as the randomization of inertial properties and of initial positions and velocities, random and adversarial disturbances, and the specification and evaluation of state and input constraints) and \emph{(ii)} the integration with approaches developed by the control theory community that leverage the knowledge of nominal models---notably, we use the symbolic framework CasADi~\cite{Andersson2019} to enrich our \texttt{gym} environments with symbolic \emph{a priori} dynamics, constraints, and cost functions.

\begin{figure}
    \centering
    \includegraphics[]{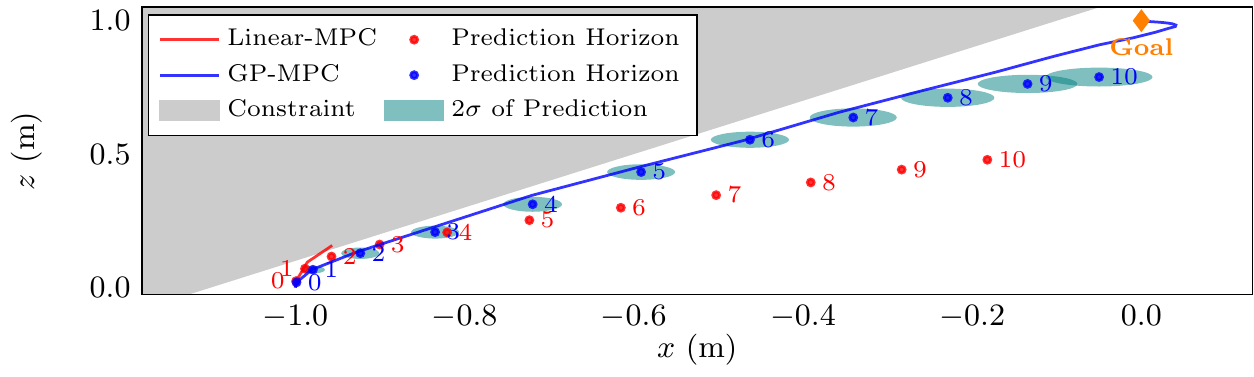}
\caption{A position comparison of the 2D quadrotor stabilization using linear MPC (red) and GP-MPC (blue), along with the prediction horizons at the second time step, subject to a diagonal state constraint (gray) and input constraints.}
\label{fig:gp-mpc}
\end{figure}

\begin{figure}
    \hspace{-1.8cm} \includegraphics[]{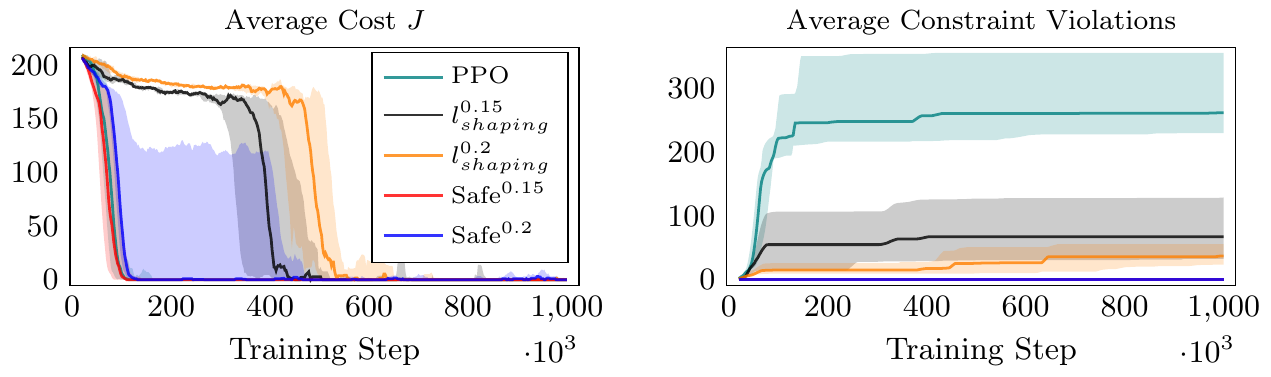}
    \vspace{3mm}
\caption{Total cost and constraint violations during learning for PPO, PPO with cost shaping, and PPO with safety layer (safe exploration). Plotted are medians with upper and lower quantiles over 10 seeds. The parameters in superscript represent values of the slack variable, which controls responsiveness to near constraint violations.}
\label{fig:safe-exp}
\end{figure}

\begin{figure}
         \hspace{1.3cm} \includegraphics[]{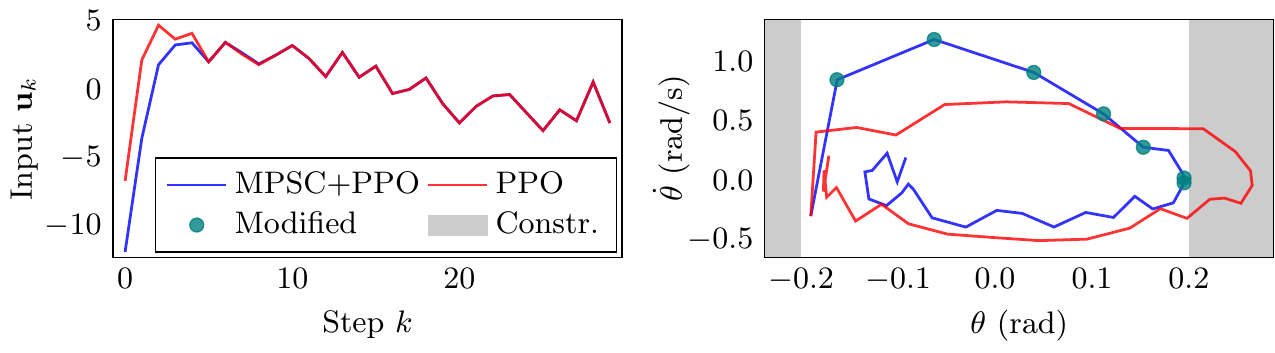}
         \vspace{3mm}
\caption{Left: The uncertified PPO input (red) is plotted against the certified MPSC+PPO input (blue).
Right: The cart-pole state diagram ($\theta$ and $\dot{\theta}$) comparing the MPSC+PPO certified trajectory (blue) and the uncertified PPO trajectory (red). The MPSC is most active (green dots) when the system is about to leave the constraint boundary (gray) or the set of states from which the MPSC can correct the system.}
\label{fig:mpc-cert}
\end{figure}

\subheading{Safe Learning Control Results.}
We focus on a constrained stabilization task, in each of our environments (\texttt{cartpole} and \texttt{quadrotor}).
The results we present here are not meant to establish the superiority of one approach over another.
Instead, we show how the approaches in \cite{Hewing2019,dalal2018a,Wabersich2018a}, taken from each of \autoref{sec:uncertain-dyn} (learning uncertain dynamics), \autoref{sec:methods-rl} (encouraging safety in RL), and \autoref{sec:certification} (certifying learning-based control), can improve control performance while pursuing constraint satisfaction.
By doing so, we demonstrate that our benchmark can be easily integrated with learnable policies developed by either the control or RL research community.
This also allows us to fairly compare the data hungriness of the different safe learning control approaches.

A learning-based robust MPC with a GP estimate of $\hat{\vec{f}}$ (GP-MPC), representative of \autoref{sec:uncertain-dyn} (learning uncertain dynamics), was implemented to stabilize a quadrotor subject to a state and input constraints, following the approach in \cite{Hewing2019}.
For this experiment, a linearization about hover with a mass and moment of inertia 150\% of the true values was used as the prior model. 
Eight-hundred randomly selected state-action pairs (or 80 seconds), sampled from a couple of minutes of training data, were used for hyper-parameter optimization, which was performed offline. 
\textbf{\autoref{fig:gp-mpc}} compares the performance of a linear MPC, using the inaccurate prior model, with the GP-MPC approach. We see that the Linear MPC, using the heavier prior model, predicts the trajectory of the quadrotor will be relatively shallow when maximum thrust is applied. This results in the quadrotor quickly violating the position constraint. In contrast, the GP-MPC is able to account for the inaccurate prior model and respects the constraint boundary by a margin proportional to the 95\% confidence interval on its dynamics prediction, stabilizing the quadrotor.

We also applied the Safe Exploration approach~\cite{dalal2018a} to the popular deep RL algorithm PPO~\cite{schulman2017proximal}, as a representative of 
\autoref{sec:methods-rl} (encouraging safety in RL), and tested it on the cart-pole stabilization task with constraints on the cart position. Notably, the task terminates upon any constraint violation. We compared it against two baselines, standard PPO and PPO with naive cost shaping (adding a penalty when close to constraint violation).
Each RL approach used over 9 hours of simulation time collecting data for training.
\textbf{\autoref{fig:safe-exp}} shows that the two constraint-aware approaches (cost shaping and safe exploration~\cite{dalal2018a}) achieve the same performance after learning, but with substantially lower constraint violations than the standard PPO. The safe exploration approach outperforms cost shaping in terms of constraint satisfaction, while not compromising convergence speed towards reaching the optimal cost. Safe Exploration, however, requires careful parameter tuning, especially on the slack variable that determines its responsiveness to near constraint violation.

Finally, a Model Predictive Safety Certification (MPSC) algorithm, based on~\cite{Wabersich2018a}, was chosen as a representative of \autoref{sec:certification} (certifying learning-based control). This particular formulation uses an MPC framework to modify an unsafe learning controller's actions. Here, a sub-optimal PPO controller provides the uncertified inputs trying to stabilize the cart-pole. The advantages of using MPSC are highlighted in~\textbf{\autoref{fig:mpc-cert}}. In \textbf{\autoref{fig:mpc-cert}} (right), the inputs are modified by the MPSC early in the stabilization to keep the cart-pole within the constraint boundaries. \textbf{\autoref{fig:mpc-cert}} (left) shows that, without the MPSC, PPO violates the constraints, but with MPSC, it manages to stay within the boundaries. The plot also shows that MPSC is most active when the system is close to the constraint boundaries (the green dots show when MPSC modified the learning controller's input). This provides a proof of concept of how safety filters can be combined with RL control for safer performance.

In our future work, we intend to use our \texttt{cartpole} and \texttt{quadrotor} benchmark environments to evaluate the robustness, performance, safety and data efficiency of different control and learning approaches.

\section{DISCUSSION AND PERSPECTIVES FOR FUTURE DIRECTIONS}
\label{sec:discussion_future}
The problem of safe learning control is emerging as a crucial topic for next-generation robotics.
In this review, we summarized approaches from the control and the machine learning communities that allow data to be safely used to improve the closed-loop performance of robot control systems.
We show that machine learning techniques, particularly RL, can help generalization towards larger classes of systems (i.e., fewer prior model assumptions), while control theory provides the insights and frameworks necessary to provide constraint satisfaction guarantees and closed-loop stability guarantees during the learning.
Despite the many advances to date, there remain many opportunities for future research.

\begin{issues}[OPEN CHALLENGES]\begin{enumerate}
\item \textbf{Capturing a Broader Class of Systems.} Work to date has focused on nonlinear systems in the form of~\autoref{eq:system_model}. 
While they can model many robotic platforms, robots can also exhibit hybrid dynamics (e.g., legged robots or other  contact dynamics with the environment~\citep{Wieber2016}), time-varying dynamics (e.g., operation in changing environments~\citep{mckinnon2018,chandak2020a}), time delays (e.g., in actuation, sensing, or observing the reward~\citep{dulac2019a}), partial differential dependencies in the dynamics (e.g., in continuum robotics~\citep{burgner-kahrs2015}). Expanding  safe learning control approaches to these scenarios is essential for their broader applicability in robotics~(see \textbf{\autoref{fig:survey_summary}}). 
\item \textbf{Accounting for Imperfect State Measurements.} The majority of safe learning control approaches assume direct access to (possibly noisy) state measurements and neglect the problem of state estimation. In practice, obtaining accurate state information is challenging due to sensors that do not provide state measurements directly (e.g., images as measurements), inaccurate process and observation models used for state estimation, and/or improper state feature representations. One open challenge is to account for state estimation errors and learned process and observation models in safe learning control~\cite{dean2020a}. Expanding existing approaches to work with (possibly high-dimensional) sensor data is essential for a broad applicability of these methods in robotics.
\item \textbf{Considering 
Scalability, and Sampling and Computational Efficiency.} 
Many of the approaches presented here have only been demonstrated on small toy problems and applying them to high-dimensional robotics problems is not trivial.  Moreover, in practice, we often face issues such as data sparsity, distribution shifts, and the optimality-complexity trade-off for real-time implementations.
Efficient robot learning relies on multiple factors including control architecture design~\citep{mueller2012iros}, systematic training data collection~\citep{Dean2018}, and appropriate function class selection~\citep{mckinnon2019b}.
 While current approaches focus on providing theoretical safety guarantees, formal analysis of sampling complexity and computational complexity 
is indispensable to facilitate the implementation of safe learning control algorithms in real-world robot applications.
\item \textbf{Verifying System and Modeling Assumptions.} The safety guarantees provided often rely on a set of assumptions (e.g., Lipschitz continuous \textit{true} dynamics with a known Lipschitz constant or bounded disturbance sets). It is difficult to verify these assumptions prior to a robot's operation. 
To facilitate algorithm implementation, we also see other approximations being made (e.g., linearization, data assumed to be independent and identically distributed (i.i.d.) Gaussian samples).  Systematic approaches to verify or quantify the impact of the assumptions and the approximations with minimal (online) data are crucial to allow the safe learning approaches to be applied in real-world applications. This can also include investigations into the interpretability of trained models, especially black-box models such as DNNs, for safe closed-loop operation~\citep{dulac2019a}.\end{enumerate}\end{issues}This list of challenges is by no means complete. Other important and open questions pertinent to safe learning control are:
\begin{itemize}
 \item What are the appropriate \textbf{benchmarks, evaluation metrics, and practices} to provide practical insights and perform fair comparisons of algorithms that rely on different assumptions?
    \item What should be the \textbf{role of simulation} in the offline design and evaluation phases? Can simulation be used to find safe hyperparameters?
    \item How do we define safety in \textbf{human-robot interaction}?
    \item How can data be safely used in \textbf{multi-agent learning} settings?
\end{itemize}

Efforts that combine control theory and machine learning for safe learning control have been shown  to result in improved control performance and system safety. Recent successes and growing interest should motivate the further development of a systematic
body of theory, advanced methodologies, and computational methods for safe learning in robotics, bringing the large potential of learning into safety-critical control and robotics applications. The open challenges and questions are intended to push us further, towards a future of real-world safe autonomy where robots reliably and safely function in complex environments.

\section*{ACKNOWLEDGMENTS}
The authors would like to acknowledge the early contributions to this work by Karime Pereida and Sepehr Samavi, the invaluable suggestions and feedback by Hallie Siegel, as well as the support from the Natural Sciences and Engineering Research Council of Canada (NSERC), the Canada Research Chairs Program, and the CIFAR AI Chair.

\appendix
\section{SUMMARY  OF REVIEWED LITERATURE}
\label{appendix:appendix}

\textbf{\autoref{tab:summary}} summarizes the safe learning control  approaches reviewed in this paper. Learning approaches are classified as \textit{Model-Based} if a dynamics model is used to produce control inputs or \textit{Model-Free} if there is a direct mapping from states or measurements to control inputs.
For model-based approaches, we  categorize the structure of the \textit{A Priori} and the \textit{Learned} component of the model as Gaussian process (\textit{GP}), neural network (\textit{NN}), or \textit{Other} for non-standard methods (e.g., set computation from data).

The \textit{Safety Properties} category highlights which algorithms can handle state and/or input constraints (\textit{Constraint Satisfaction}) and which ones provide stability guarantees (\textit{Stability}).
For both of these properties, we report whether they are enforced while the policy is being trained (\textit{During Learning}) or by the resultant policy (\textit{After Learning}).
The \textit{Robustness} category indicates whether the final policy is designed to be robust to \textit{Input} disturbances; model \textit{Parameter} uncertainty, and/or other \textit{Dynamics} uncertainties broader than the previous two types of disturbances.

Finally, the last category reports the \textit{Task(s)} to which each method was applied: \textit{Stabilization} for stabilizing an equilibrium of the system; \textit{Tracking} for the tracking of a given trajectory; \textit{Navigation} for navigating or planning a sequence of actions to reach a given goal; \textit{Locomotion} for the movement of legged robots like humanoids or quadrupeds; and \textit{Manipulation} for pushing, reaching, or grasping operations using a robotic arm.

\begin{landscape}
\begin{table}[ht!]
\tabcolsep7.5pt
    \caption{
    Summary of the key properties of the works
referenced in \autoref{sec:methods}
    }
    \label{tab:summary}
    \begin{center}
    \resizebox{1.1\columnwidth}{!}{
    { \begin{tabular}{p{1.5cm} p{0.5cm} p{2.2cm} p{2cm} p{1.6cm} p{1.9cm} p{1.5cm} p{2.1cm} p{1.5cm} p{2.1cm} p{4cm}}

   \toprule

   \multicolumn{2}{c}{\multirow{4}{*}{Section}}
    & \multicolumn{3}{c}{Learning}
    & \multicolumn{5}{c}{Safety Properties}
    & \multicolumn{1}{c}{\multirow{4}{*}{Task}}
   \\
    
   \cmidrule(lr){3-5}
   \cmidrule(lr){6-10}
    
& & \multicolumn{2}{c}{Model-based}
    & \multicolumn{1}{c}{\multirow{3}{*}{Model-free}}
    & \multicolumn{2}{c}{During Learning}
    & \multicolumn{3}{c}{After Learning}
    & \\
    
   \cmidrule(lr){3-4}
   \cmidrule(lr){6-7}
   \cmidrule(lr){8-10}
    
& & \multicolumn{1}{c}{\multirow{2}{*}{\emph{A Priori}}}
    & \multicolumn{1}{c}{\multirow{2}{*}{Learned}}
    & & \multicolumn{1}{c}{Constraint}
    & \multicolumn{1}{c}{\multirow{2}{*}{Stability}}
    & \multicolumn{1}{c}{Constraint}
    & \multicolumn{1}{c}{\multirow{2}{*}{Stability}}
    & \multicolumn{1}{c}{Robustness}
    & \\
    
& & 
    & 
    & & \multicolumn{1}{c}{Satisfaction}
    & 
    & \multicolumn{1}{c}{Satisfaction}
    & 
    & \multicolumn{1}{c}{w.r.t.}
    & \\
    
   \cmidrule(lr){1-2}
   \cmidrule(lr){3-3}
   \cmidrule(lr){4-4}
   \cmidrule(lr){5-5} 
   \cmidrule(lr){6-6}
   \cmidrule(lr){7-7}
   \cmidrule(lr){8-8}
   \cmidrule(lr){9-9}
   \cmidrule(lr){10-10}
   \cmidrule(lr){11-11}

    \multirow{15}{*}{\shortstack{Learning\\Uncertain\\Dynamics\\\autoref{sec:uncertain-dyn}}}
    & Sec. \ref{sec:learning_adaptation}
    & Linear~\cite{cooper2014use, Gahlawat2020,joshi2019deep, joshi2020asynchronous}, Nonlinear~\cite{grande2014experimental,chowdhary2014bayesian} &  NN~\cite{cooper2014use,joshi2019deep, joshi2020asynchronous}, GP~\cite{Gahlawat2020,grande2014experimental, chowdhary2014bayesian} & \cellcolor{black!10} & \cellcolor{black!10} & \cite{cooper2014use, Gahlawat2020,grande2014experimental,chowdhary2014bayesian,joshi2019deep, joshi2020asynchronous} & \cellcolor{black!10} & \cite{cooper2014use, Gahlawat2020,grande2014experimental,chowdhary2014bayesian,joshi2019deep, joshi2020asynchronous} & Input~\cite{cooper2014use, Gahlawat2020}, Dynamics~\cite{grande2014experimental,chowdhary2014bayesian,joshi2019deep, joshi2020asynchronous} & Tracking~\cite{cooper2014use, Gahlawat2020,grande2014experimental,chowdhary2014bayesian,joshi2019deep, joshi2020asynchronous} \\
    
   \cmidrule(lr){2-11}

   & Sec. \ref{sec:learning_robust_control}
    &  Linear~\cite{Berkenkamp2015a, holicki2021a, vonrohr2021probabilistic}, Nonlinear~\cite{Helwa2019, greeff2021a} & GP~\cite{Berkenkamp2015a, vonrohr2021probabilistic, Helwa2019, greeff2021a} Other~\cite{holicki2021a} & \cellcolor{black!10} & \cellcolor{black!10} & \cite{Berkenkamp2015a, vonrohr2021probabilistic, Helwa2019, greeff2021a} & \cellcolor{black!10} & \cite{Berkenkamp2015a,  holicki2021a, vonrohr2021probabilistic, Helwa2019, greeff2021a} & Dynamics~\cite{Berkenkamp2015a,  holicki2021a, vonrohr2021probabilistic, Helwa2019, greeff2021a} & Stabilization~\cite{Berkenkamp2015a,  holicki2021a, vonrohr2021probabilistic}, Tracking~\cite{Helwa2019, greeff2021a}   \\
    
   \cmidrule(lr){2-11}

   & Sec. \ref{sec:learning_mpc}
    & Linear~\cite{Tanaskovic2014,Lorenzen2019, Bujarbaruah2018, Bujarbaruah2019, Bujarbaruah2018c, Pereida2021,Aswani2013, Soloperto2018}, Nonlinear~\cite{Goncalves2016, Kohler2020,Ostafew2016a,Hewing2019,kamthe2018a,Koller2019b,Fan2020,mckinnon2020a} & NN~\cite{Aswani2013,Fan2020}, GP~\cite{Soloperto2018,Ostafew2016a,Hewing2019,kamthe2018a,Koller2019b}, Other~\cite{Tanaskovic2014,Lorenzen2019, Bujarbaruah2018, Bujarbaruah2019,Goncalves2016,Kohler2020, Bujarbaruah2018c,Pereida2021, mckinnon2020a} & \cellcolor{black!10} & Input {\&} state \cite{Tanaskovic2014,Lorenzen2019, Bujarbaruah2018, Bujarbaruah2019, Goncalves2016, Kohler2020, Bujarbaruah2018c, Pereida2021, Aswani2013, Soloperto2018,Ostafew2016a,Hewing2019,kamthe2018a,Koller2019b,Fan2020}& \cite{Tanaskovic2014,Lorenzen2019, Bujarbaruah2018, Bujarbaruah2019,Goncalves2016, Kohler2020, Bujarbaruah2018c, Pereida2021, Aswani2013, Soloperto2018,Ostafew2016a,Hewing2019,kamthe2018a,Koller2019b,Fan2020} & Input {\&} state \cite{Tanaskovic2014,Lorenzen2019, Bujarbaruah2018, Bujarbaruah2019,Goncalves2016, Kohler2020, Bujarbaruah2018c, Pereida2021, Aswani2013, Soloperto2018,Ostafew2016a,Hewing2019,kamthe2018a,Koller2019b,Fan2020,mckinnon2020a}& \cite{Tanaskovic2014,Lorenzen2019, Bujarbaruah2018, Bujarbaruah2019, Goncalves2016, Kohler2020, Bujarbaruah2018c, Pereida2021, Aswani2013, Soloperto2018,Ostafew2016a,Hewing2019,kamthe2018a,Koller2019b,Fan2020,mckinnon2020a} & Input~\cite{Pereida2021}, Dynamics~\cite{Aswani2013, Soloperto2018,Ostafew2016a,Hewing2019,kamthe2018a,Koller2019b,Fan2020,mckinnon2020a}, Parameters~\cite{Tanaskovic2014,Lorenzen2019, Bujarbaruah2018, Bujarbaruah2019,Goncalves2016, Kohler2020, Bujarbaruah2018c, Pereida2021} & Stabilization~\cite{Lorenzen2019, Bujarbaruah2018, Bujarbaruah2019,Goncalves2016,Kohler2020, Bujarbaruah2018c,Pereida2021,  Aswani2013,Soloperto2018, kamthe2018a, Koller2019b}, Tracking~\cite{Tanaskovic2014,Ostafew2016a,Hewing2019,Fan2020, mckinnon2020a} \\
    
   \cmidrule(lr){2-11}

   & Sec. \ref{sec:smbrl}
    &  Nonlinear~\cite{Berkenkamp2017} & GP~\cite{Berkenkamp2017} & \cellcolor{black!10} & \cellcolor{black!10} & \cite{Berkenkamp2017} & \cellcolor{black!10} & \cite{Berkenkamp2017} & \cellcolor{black!10} & Stabilization~\cite{Berkenkamp2017} \\
    
   \cmidrule(lr){1-11}

    \multirow{18}{*}{\shortstack{Encouraging\\Safety\\in RL\\\autoref{sec:methods-rl}}}
    & Sec. \ref{sec:methods-rl:safe-exp}
    & Nonlinear~\cite{Turchetta2016,Berkenkamp2016a, wachi2018a}, & GP~\cite{sui2015a,Berkenkamp2016a, sui2018a,baumann2021gosafe} & NN~\cite{dalal2018a,pham2018optlayer,srinivasan2020learning,thananjeyan2020b,bharadhwaj2021conservative}, Tabular~\cite{moldovan2012a} & Input {\&} state \cite{Turchetta2016,dalal2018a,moldovan2012a,pham2018optlayer,sui2015a,Berkenkamp2016a,sui2018a,baumann2021gosafe,wachi2018a}
    Trajectory~\cite{bharadhwaj2021conservative} & \cellcolor{black!10} & Input {\&} state \cite{Turchetta2016,dalal2018a,moldovan2012a,pham2018optlayer,sui2015a,Berkenkamp2016a,sui2018a,baumann2021gosafe,wachi2018a}
    Trajectory~\cite{srinivasan2020learning,thananjeyan2020b,bharadhwaj2021conservative} & \cellcolor{black!10} & \cellcolor{black!10} & Stabilization~\cite{baumann2021gosafe}, Tracking~\cite{Berkenkamp2016a}, Navigation~\cite{Turchetta2016,dalal2018a,moldovan2012a,wachi2018a,srinivasan2020learning,thananjeyan2020b}, Locomotion~\cite{srinivasan2020learning,bharadhwaj2021conservative}, Manipulation~\cite{pham2018optlayer,srinivasan2020learning,thananjeyan2020b,bharadhwaj2021conservative} \\
    
   \cmidrule(lr){2-11}

   & Sec. \ref{sec:methods-rl:risk}
    & Nonlinear~\cite{kahn2017a,lutjens2019a} & NN~\cite{kahn2017a,lutjens2019a,zhang2020cautious,thananjeyan2020a} & NN~\cite{urpi2021riskaverse} & \cellcolor{black!10} & \cellcolor{black!10} & Input {\&} state \cite{kahn2017a,lutjens2019a,zhang2020cautious,thananjeyan2020a} & \cellcolor{black!10} & Dynamics~\cite{urpi2021riskaverse} & Stabilization~\cite{zhang2020cautious}, Navigation~\cite{kahn2017a,lutjens2019a,zhang2020cautious,thananjeyan2020a},  Locomotion~\cite{zhang2020cautious,urpi2021riskaverse}, Manipulation~\cite{thananjeyan2020a} \\
    
   \cmidrule(lr){2-11}

   & Sec. \ref{sec:methods-rl:cmdps}
    & Tabular~\cite{chow2018lyapunov, satija2020constrained}
    & \cellcolor{black!10}
    & NN~\cite{achiam2017constrained,liang2018accelerated, chow2018lyapunov, chow2019lyapunov, satija2020constrained}
    & Trajectory~\cite{achiam2017constrained, chow2018lyapunov, chow2019lyapunov, satija2020constrained}
    & \cellcolor{black!10}
    & Trajectory~\cite{achiam2017constrained,liang2018accelerated, chow2018lyapunov, chow2019lyapunov, satija2020constrained}
    & \cellcolor{black!10}
    & \cellcolor{black!10}
    & Navigation~\cite{achiam2017constrained,liang2018accelerated, chow2018lyapunov, chow2019lyapunov, satija2020constrained}, Locomotion~\cite{achiam2017constrained, chow2019lyapunov, satija2020constrained}
   \\
    
   \cmidrule(lr){2-11}

   & Sec. \ref{sec:methods-rl:rrl}
    & \cellcolor{black!10}
    & NN~\cite{loquercio2019a}
    & NN~\cite{pinto2017robust, pan2019a, vinitsky2020robust, everett2020certified, sadeghi2017card2rl, rajeswaran2017epopt, mehta20a}
    & Input {\&} state \cite{loquercio2019a}
    & \cite{loquercio2019a}
    & Input {\&} state \cite{loquercio2019a}
    & \cite{loquercio2019a}
    & Input~\cite{pan2019a, vinitsky2020robust}, Dynamics~\cite{pinto2017robust, everett2020certified, sadeghi2017card2rl, loquercio2019a}, Parameters~\cite{rajeswaran2017epopt, mehta20a}
    & Stabilization~\cite{pinto2017robust, everett2020certified, mehta20a}, Tracking~\cite{loquercio2019a}, Navigation~\cite{everett2020certified, sadeghi2017card2rl}, Locomotion~\cite{pinto2017robust, pan2019a, vinitsky2020robust, rajeswaran2017epopt}, Manipulation~\cite{mehta20a}
   \\
    
   \cmidrule(lr){1-11}

    \multirow{13}{*}{\shortstack{Certifying\\Learning-\\based\\Control\\\autoref{sec:certification}}}
    & Sec. \ref{sec:stability_certification}
    & Linear ~\cite{jin2020stability}, Nonlinear~\cite{zhou2020b,shi2019a, Richards2018a,zhou2020general} & NN~\cite{zhou2020b,jin2020stability,shi2019a,Richards2018a}, Other~\cite{zhou2020general} & \cellcolor{black!10} & \cellcolor{black!10} & \cite{jin2020stability} & \cellcolor{black!10} & \cite{zhou2020b,jin2020stability,shi2019a,Richards2018a,zhou2020general} & Dynamics~\cite{zhou2020b,jin2020stability,shi2019a, Richards2018a,zhou2020general}  & Stabilization~\cite{Richards2018a,zhou2020general}, Tracking~\cite{zhou2020b,jin2020stability,shi2019a,zhou2020general}, Locomotion~\cite{zhou2020general} \\
    
   \cmidrule(lr){2-11}

    & Sec. \ref{sec:set_certification}
    & Linear~\cite{Wabersich2018a,Wabersich2019}, Nonlinear~\cite{taylor2019a,taylor2020a,Ohnishi2019, Choi2020,taylor2019b, taylor2020b,taylor2020d,taylor2020c,Lopez2020,cheng2019a,Fan2019,Wang2018a,Fisac2019,Gillula2012,bajcsy19efficient,Wabersich2021,mannucci2018a}
    & GP~\cite{Ohnishi2019,cheng2019a,Wang2018a,khojasteh2020a,Fisac2019,Wabersich2019}, NN~\cite{taylor2019a,taylor2020a,Choi2020,taylor2019b,taylor2020b,cheng2019a,Fan2019,bajcsy19efficient}
    Other~\cite{taylor2020d,taylor2020c,Lopez2020,Gillula2012,Wabersich2018a,Wabersich2019,Wabersich2021}
& NN~\cite{Ohnishi2019,Choi2020, Fisac2019a} & Input {\&} state \cite{taylor2019b, taylor2020b,taylor2020d,taylor2020c,Lopez2020,cheng2019a,Fan2019,Wang2018a,Wabersich2018a,Wabersich2019,Wabersich2021,mannucci2018a}, State~\cite{Fisac2019,Gillula2012,bajcsy19efficient}
    & \cite{taylor2019b, taylor2020b,taylor2020d,taylor2020c,Lopez2020,cheng2019a,Fan2019,Wang2018a,Fisac2019,Gillula2012,bajcsy19efficient,Wabersich2018a,Wabersich2019,Wabersich2021}
    & Input {\&} state \cite{taylor2019a,taylor2020a,Ohnishi2019, Choi2020,taylor2019b, taylor2020b,taylor2020d,taylor2020c,Lopez2020,cheng2019a,Fan2019,Wang2018a,khojasteh2020a,Wabersich2018a,Wabersich2019,Wabersich2021,mannucci2018a}, State~\cite{Fisac2019,Gillula2012,bajcsy19efficient,Fisac2019a}
    & \cite{taylor2019a,taylor2020a,Ohnishi2019,Choi2020,taylor2019b, taylor2020b,taylor2020d,taylor2020c,Lopez2020,cheng2019a,Fan2019,Wang2018a,khojasteh2020a,Fisac2019,Gillula2012,bajcsy19efficient,Fisac2019a,Wabersich2018a,Wabersich2019,Wabersich2021}
    & Dynamics~\cite{taylor2019a,taylor2020a,Ohnishi2019, Choi2020,taylor2019b, taylor2020b,taylor2020d,cheng2019a,Fan2019,Wang2018a,Fisac2019,Gillula2012,bajcsy19efficient,Fisac2019a,Wabersich2018a,Wabersich2019,mannucci2018a}, Parameters~\cite{taylor2020c,Lopez2020,Wabersich2021}
    & Stabilization~\cite{cheng2019a,Fisac2019a,Wabersich2018a,Wabersich2021,mannucci2018a}, Tracking~\cite{taylor2019a,taylor2020a,taylor2019b, taylor2020b,taylor2020d,taylor2020c,Lopez2020,Fan2019,Wang2018a,khojasteh2020a,Fisac2019,Gillula2012,Wabersich2019},
    Navigation~\cite{Ohnishi2019,bajcsy19efficient},
    Locomotion~\cite{Choi2020}
   \\

\bottomrule
\end{tabular}
}

     }
    \end{center}
\end{table}
\end{landscape}

%\appendix
\section{REVISION HISTORY}
\label{appendix:appendix-b}

Send errata to \href{mailto:angela.schoellig@robotics.utias.utoronto.ca}{\texttt{angela.schoellig@robotics.utias.utoronto.ca}}.

\hfill\\

\begin{tabular}{ l p{6cm} }
   \toprule
   \bfseries Date
   & \multicolumn{1}{l}{\bfseries Change} \\
   
    \cmidrule(lr){1-2}
    
   December 6, 2021
   & Figure 6 update \\
   
    \cmidrule(lr){1-2}
    
   August 13, 2021
   & Initial submission \\
   
   \bottomrule

\end{tabular}

\end{document}